\documentclass[10pt,twocolumn,letterpaper]{article}

\usepackage[pagenumbers]{iccv} 

\definecolor{iccvblue}{rgb}{0.21,0.49,0.74}
\usepackage[pagebackref,breaklinks,colorlinks,allcolors=iccvblue]{hyperref}
\usepackage[T1]{fontenc}
\usepackage{float}
\usepackage{multirow}
\usepackage{booktabs}
\usepackage{caption}
\usepackage{graphicx}
\usepackage{amsmath, bm}
\usepackage{lipsum}
\usepackage{microtype}
\usepackage{mathrsfs}

\usepackage{multirow}
\usepackage{booktabs}
\usepackage{graphicx}
\usepackage{amsmath, bm}
\usepackage{color}
\usepackage{colortbl}  
\usepackage{xcolor}
\usepackage{ulem}
\usepackage{listings}
\usepackage{xcolor}
\usepackage{float}
\usepackage{siunitx}
\usepackage{mathrsfs}
\usepackage{fancyhdr}
\def\paperID{7851} 
\def\confName{ICCV}
\def\confYear{2025}
\fancypagestyle{firstpage}{
  \fancyhf{} 
  \fancyfoot[L]{\rule[0pt]{0.49\textwidth}{0.4pt}\\$*$ Corresponding Author\\ \textit{Preprint }}

}
\title{S2CFormer: Revisiting the RD-Latency Trade-off in Transformer-based Learned Image Compression}

\author{
\centering
\begin{tabular}{c}
   Yunuo Chen\textsuperscript{$\dagger$}\quad
   Qian Li\textsuperscript{$\dagger$}\quad
   Bing He\textsuperscript{$\dagger$}\quad
   Donghui Feng\textsuperscript{$\dagger$}\quad
   Ronghua Wu\textsuperscript{$\ddagger$}\quad
   Qi Wang\textsuperscript{$\ddagger$}\\
   Li Song\textsuperscript{$\dagger$}\quad
   Guo Lu\textsuperscript{$\dagger$}* \quad
   Wenjun Zhang\textsuperscript{$\dagger$}\\
   \textsuperscript{$\dagger$}Shanghai Jiao Tong University\quad
   \textsuperscript{$\ddagger$}Ant Group
\end{tabular}
}

\begin{document}
\twocolumn[{%
\maketitle
\thispagestyle{firstpage} 
\begin{figure}[H]
\hsize=\textwidth 
\centering
\includegraphics[width=17.5cm]{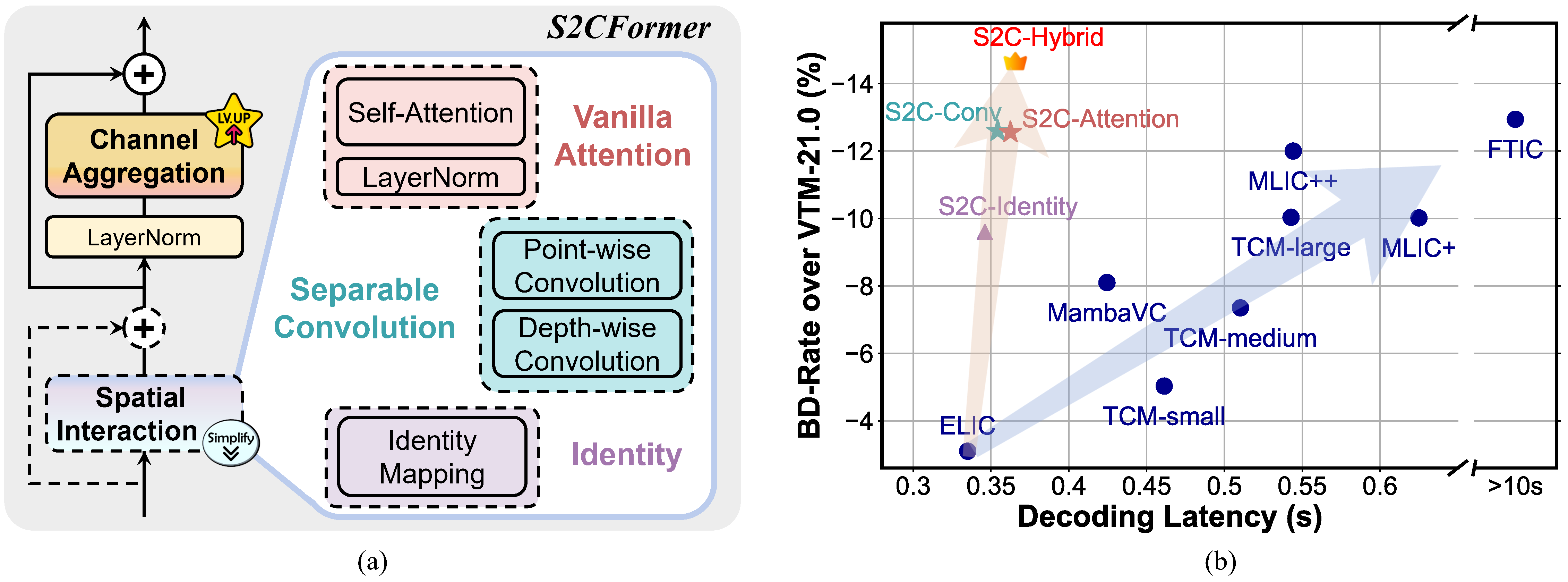}
\caption{\textbf{S2CFormer and the performance of S2CFormer-based models.} The general structure of our S2CFormer is shown in (a). It consists of two key components: the Spatial Interaction module and the Channel Aggregation module. S2CFormer functions as nonlinear transform blocks for Learned Image Compression (LIC). Our analysis reveals that the competence of transformer-based LIC models primarily stems from channel aggregation.
Building on this insight, we propose a novel design strategy that rebalances these two modules to achieve a more favorable trade-off between compression performance and decoding latency.  As illustrated in (b),   the data points for S2CFormer-based models exhibit a linear trend with a steeper slope, thereby underscoring their superior performance–latency characteristics. 
}
\label{start}
\end{figure}
}]

\begin{abstract}
Transformer-based Learned Image Compression (LIC) suffers from a suboptimal trade-off between decoding latency and rate-distortion (R-D) performance. Moreover, the critical role of the FeedForward Network (FFN)-based channel aggregation module has been largely overlooked. Our research reveals that efficient channel aggregation—rather than complex and time-consuming spatial operations—is the key to achieving competitive LIC models.
Based on this insight, we initiate the ``S2CFormer'' paradigm, a general architecture that simplifies spatial operations and enhances channel operations to overcome the previous trade-off. 
We present two instances of the S2CFormer: S2C-Conv, and S2C-Attention.  Both models demonstrate state-of-the-art (SOTA) R-D performance and significantly faster decoding speed, as shown in Fig. \ref{start}. 
Furthermore, we introduce S2C-Hybrid, an enhanced variant that maximizes the strengths of different S2CFormer instances to achieve a better performance-latency trade-off. This model outperforms all the existing methods on the Kodak, Tecnick, and CLIC Professional Validation datasets, setting a new benchmark for efficient and high-performance LIC. 
The code is at \href{https://github.com/YunuoChen/S2CFormer}{https://github.com/YunuoChen/S2CFormer}.

\end{abstract}
\section{Introduction}
\label{sec:intro} 
\begin{figure}
\centering
\includegraphics[width=0.47\textwidth]{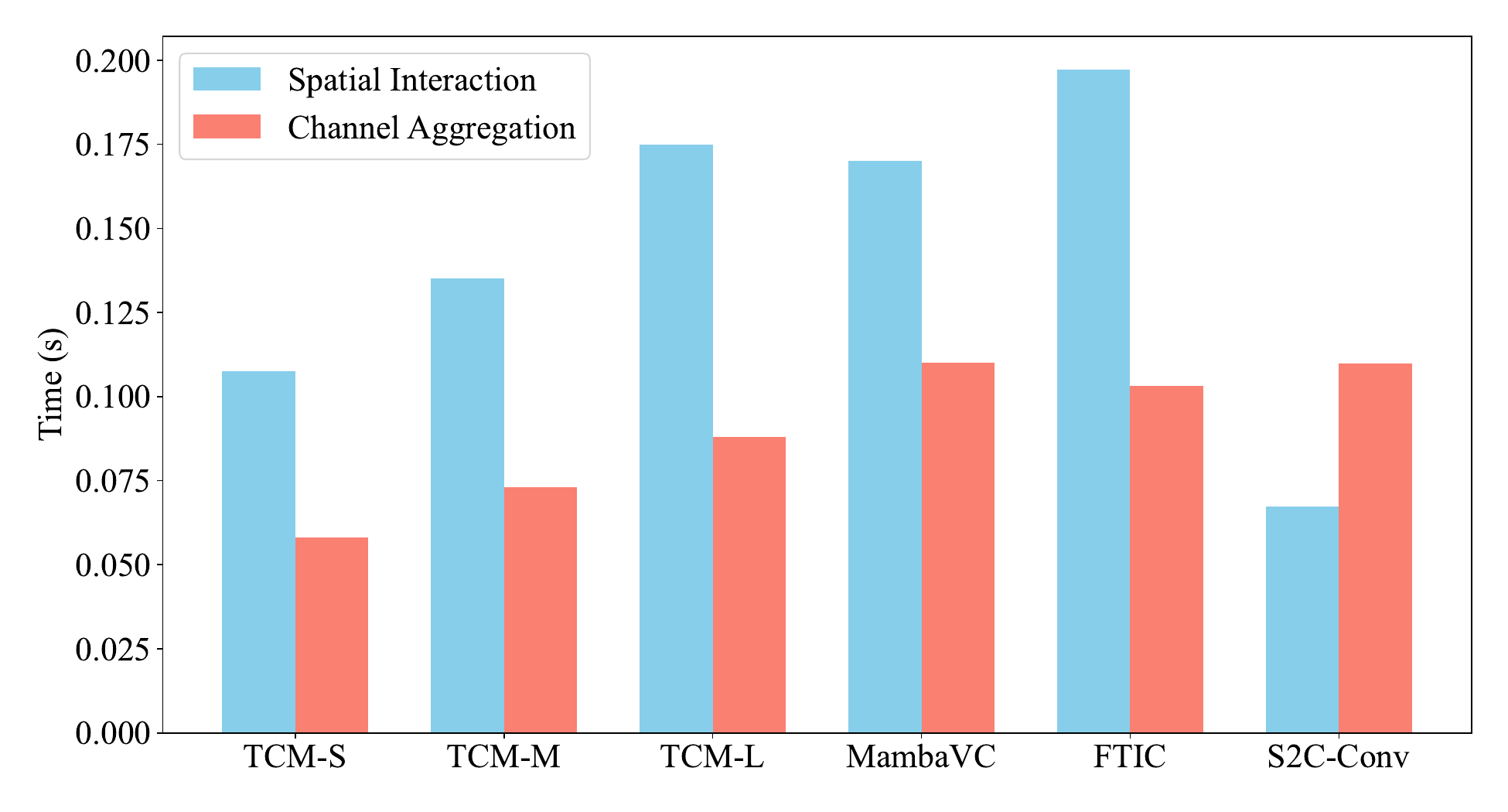}
\captionsetup{skip=1pt}  
\caption{\textbf{Comparison of execution times for spatial interaction and channel aggregation across different models.} Previous methods show much higher spatial interaction times than channel aggregation, causing significant delays. Our S2CFormer effectively rebalances the time relationship between these two modules.
}
\vspace{-10pt}  
\label{intro} 
\end{figure}
Image compression is crucial due to the rapid growth of digital image data. Lossy compression reduces file sizes while preserving visual quality and optimizing storage and transmission. Recently, learned image compression (LIC) models have emerged as a promising alternative to traditional codecs, offering improved rate-distortion performance and greater efficiency in image storage.

For LIC, nonlinear transform blocks and entropy models are pivotal to a model's performance. Early LIC models primarily relied on convolutional neural networks (CNNs) for transform networks \cite{cheng2020learned, chen2022two, minnen2018joint, balle2018variational, he2022elic, minnen2020channel, xie2021enhanced, ma2020end}. However, with the advancements in transformer-based architectures \cite{vaswani2017attention, liu2021swin}, recent studies have adopted transformers as the foundation, leading to significant improvements in rate-distortion (R-D) performance \cite{zhu2022transformer, liu2023learned, li2023frequency, koyuncu2022contextformer, qian2022entroformer, ren2023bayesian, he2024s4d, zhang2023neural}. 

The success of transformers is often attributed to their sophisticated spatial operations, such as self-attention and window shifting. However, the significance of the transformer structure has not been thoroughly explored, making such attributions premature. Besides, as shown in Fig. \ref{start} and Fig. \ref{intro},  their complex and heavy spatial operations are quite time-consuming, thus often leading to a suboptimal trade-off between decoding latency and R-D performance.

To this end, we reevaluate the key factors in the R-D performance of transformers-based LIC models. We identify that a basic vision transformer block \cite{vaswani2017attention, liu2021swin, liu2022swin} consists of two operations targeting different dimensions: an attention-based spatial operation and an MLP-based channel operation. By replacing spatial operations with identity mapping, we are surprised to find that channel operations alone can achieve promising performance, which is even comparable to the leading methods. 
The solid lower performance bound and the simple identity spatial operation emphasize that the MLP-based channel operation plays an essential role in optimizing R-D performance for LIC models, while previously complex spatial interactions may be partly redundant.  
To reorient the focus of LIC from \textbf{S}patial Interaction \textbf{to} \textbf{C}hannel Aggregation, 
we introduce the paradigm ``S2CFormer''. S2CFormer is a general structure originated from the standard vision transformers, which consists of two main components: Spatial Interaction and Channel Aggregation. This paradigm emphasizes two key ideas: 1) simplifying spatial interactions to accelerate decoding, and 2) aggregating anisotropic features along the channel dimension to achieve promising performance. It realigns the priority of the spatial operator and channel operator to break the limitations of the previous speed-performance trade-off.

Based on the insight of S2CFormer, we adopt two basic operators for spatial interaction: separable convolution, and vanilla attention. These operators, combined with a FFN for channel aggregation, constitute two basic S2CFormer instances: S2C-Conv, and S2C-Attention. S2C-Conv and S2C-Attention demonstrate that combining simple spatial interaction with channel aggregation can achieve SOTA performance and improve decoding speed by over 30\%. Their performances indicate that channel aggregation is the key factor in R-D performance, far outweighing the choice of specific spatial interaction methods (e.g., convolution or attention). We attribute this to reduced feature resolution shifting spatial redundancy to channels, highlighting the importance of capturing and eliminating channel redundancy.

These findings motivate further exploration of advanced MLP structures for channel aggregation, which could enhance LIC models with minimal added complexity. 
Additionally, we propose a new LIC model, S2C-Hybrid. This model achieves even better compression performance by appropriately arranging and combining different S2C-instances, thus fully exploiting the strengths of each S2C-instantiation.
Besides, it also ensures a much faster decoding speed compared to existing SOTA methods.  

Our contributions can be summarized as follows:

\begin{itemize}
  \item [$\bullet$] Revealing the Importance of Channel Aggregation in LIC: We emphasize the critical role of channel aggregation in LIC and point out that the previous excessive focus on spatial interactions leads to a suboptimal trade-off between decoding latency and R-D performance.

  \item [$\bullet$] Introduction of the S2CFormer Paradigm: 
    We propose S2CFormer to reorient the focus of LIC from spatial interactions to channel aggregation. This paradigm advocates for simplifying spatial interaction and enhancing channel aggregation to improve both speed and performance.

  \item [$\bullet$]  Establishing New Performance Benchmarks:  
  The S2C-Hybrid model combines the strengths of different S2C-instances and outperforms all the existing methods on three datasets (i.e., Kodak, Tecnick, and CLIC datasets), establishing a new R-D performance record for LIC.
\end{itemize}

\section{Related Work}
\label{sec:Related work}
\subsection{Learned Image Compression}

In the past few years, end-to-end learned image compression (LIC) has gained increased attention. Ball\'{e} \textit{et al.} \cite{balle2016end} proposed the first end-to-end image compression model with convolution neural network (CNN) and introduced the variation auto-encoder (VAE) model combined with hyper-prior \cite{balle2018variational}, which has become the fundamental structure for LIC models. Afterward, researchers mainly focus on two things to improve the rate-distortion performance: transform networks and entropy model \cite{zafari2023frequency, mentzer2022vct, ma2019iwave, lu2022transformer, liu2020unified, gao2021neural, fu2023asymmetric, begaint2020compressai, han2024causal}.

Transform networks refer to the nonlinear analysis and synthesis transforms in Encoder, Decoder, Hyper Encoder, and Hyper Decoder. Cheng \textit{et al.} \cite{cheng2020learned} adopt Residual blocks. Chen \textit{et al.} \cite{chen2022two} optimize it with octave residual modules. Xie  \textit{et al.} \cite{xie2021enhanced} utilized an invertible neural network for better performance. With the development of various transformers, they are also adopted by LIC models.  \cite{zou2022devil, zhu2022transformer} directly utilized Swin Transformer for transforms. TCM  \cite{liu2023learned} employed both Resblock and Swin Transformer to capture both local and global information. FTIC  \cite{li2023frequency} enhances the window attention mechanism from a frequency perspective by carefully designing the attention window.

For the entropy model, researchers mainly focus on exploiting efficient and powerful context models.  Minnen \textit{et al.} \cite{minnen2018joint} proposed an autoregressive entropy model. He \textit{et al.}  \cite{he2021checkerboard} utilized the checkerboard model to improve the speed. Minnen  \textit{et al.} \cite{minnen2020channel} proposed a context model along channel dimension. He \textit{et al.} \cite{he2022elic} raised ELIC for unevenly grouped space-channel contextual adaptive coding.  \cite{li2022hybrid, li2023neural} proposed a quadtree entropy model for efficiency.
 Qian \textit{et al.} \cite{qian2022entroformer} utilized ViT for the entropy model for powerful context. MLIC and MLIC++  \cite{jiang2023mlic, jiang2023mlic++} proposes a multi-reference entropy model. FTIC  \cite{li2023frequency} propose a transformer-based channel-wise autoregressive (T-CA) model that effectively exploits channel dependencies. In this paper, we mainly focus on the structure of transform networks, and directly adopt the basic efficient and powerful entropy model from ELIC \cite{he2022elic}.

\subsection{Transformer}

Transformers, initially introduced by \cite{vaswani2017attention} for translation tasks, has quickly become widely adopted in a range of NLP applications. Building on this success, many researchers have expanded the use of attention mechanisms and transformers to tackle computer vision tasks. \cite{dosovitskiy2020image, liu2021swin, touvron2021training, zamir2022restormer, chen2022simple}. 
In particular, Liu \textit{et al.} \cite{liu2021swin} introduced the Swin Transformer, a hierarchical model that employs window attention and window shifting mechanisms. This innovative structure has established a new performance benchmark in both high-level \cite{xia2022vision, dong2022cswin, liu2022swin} and low-level  \cite{liang2021swinir, zamir2022restormer, chen2022simple, zhang2022practical} vision tasks.

Many LIC models have adopted the Swin Transformer for nonlinear transform, achieving remarkable improvements in RD performance.  
The success of those models has been long attributed to the sophisticated attention modules. However, in the field of natural language processing (NLP), Thorp \textit{et al.} suggest in their work \cite{lee2021fnet} that substituting the attention module with a simple Fourier transform can yield comparable results. Similarly, in high-level vision tasks, Yu \textit{et al.}~\cite{yu2022metaformer, yu2023metaformer} demonstrate that the architecture known as MetaFormer is actually the most crucial factor for achieving high performance.
In this paper, we revisit the competence of transformers in LIC models, arguing that the core factor for superior R-D performance is Channel Aggregation, rather than the well-designed attention modules.

\section{Methodology}
\label{sec:Methodology}
\subsection{Problem Formulation}
\label{Problem Formulation} 
\begin{figure*}
\centering
\includegraphics[width=1\textwidth]{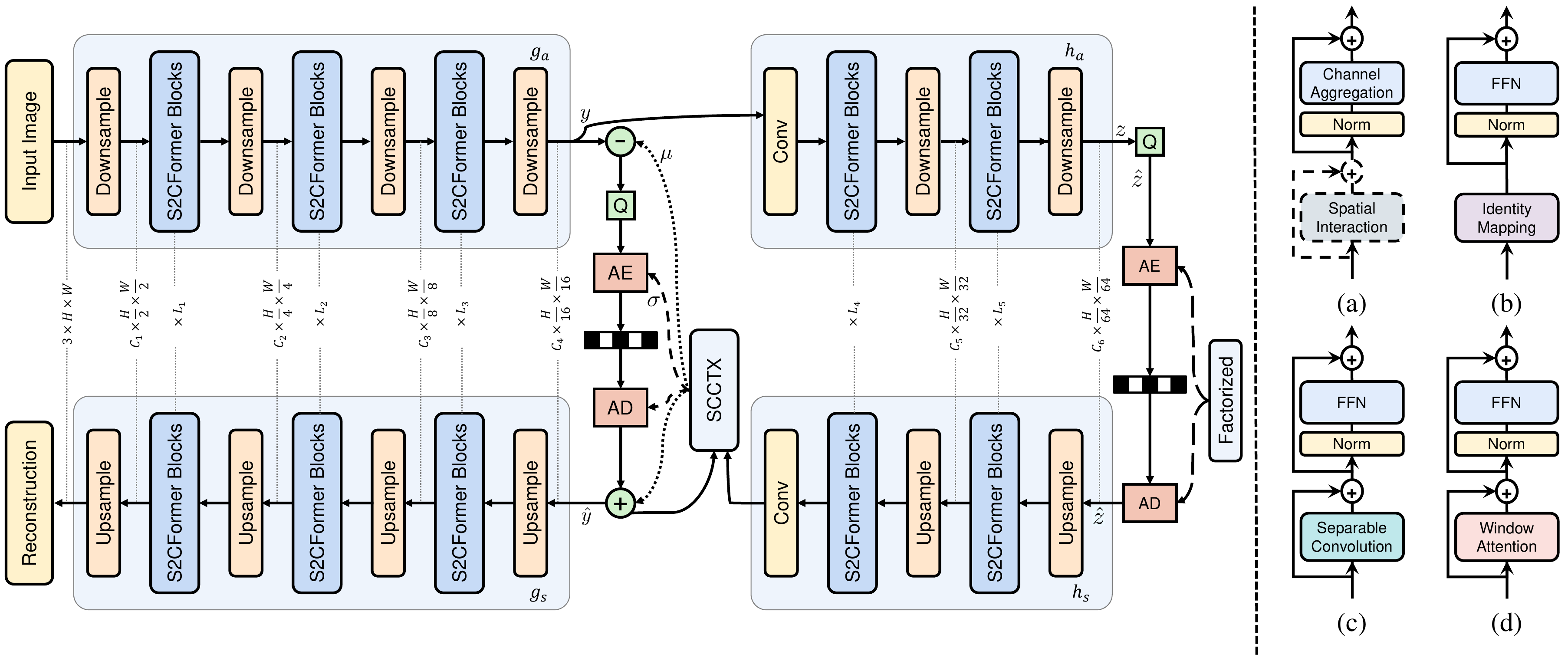}
\captionsetup{skip=1pt}  
\caption{\textbf{Overview of S2CFormer-based LIC model.} We adopt the basic VAE structure from \cite{minnen2018joint, balle2018variational} and integrate the SCCTX entropy model from \cite{he2022elic}. The hierarchical architecture consists of five stages of nonlinear transform blocks. Each stage contains \(L_i\) S2CFormer blocks. The general S2CFormer architecture is shown in (a), and (b-d) illustrate three S2CFormer instances. $L_1$-$L_6$ and $C_1$-$C_6$ represent block numbers and channel numbers for each stage, respectively}
\label{backbone} 
\end{figure*}

The structure of our S2CFormer-based LIC model is illustrated in Fig. \ref{backbone}. It consists of three main components: an Encoder, a Decoder, and an Entropy Model. The Encoder generates a latent representation \( \bm{y} \) from RGB input \( \bm{x} \), which is quantized to \( \hat{\bm{y}} \). The Decoder maps \( \hat{\bm{y}} \) back to the RGB domain. This process involves an encoder network \( g_a(\cdot) \), a decoder network \( g_s(\cdot) \), and a quantization operator \( Q \):
\begin{equation}
\boldsymbol{y} = g_a(\boldsymbol{x}; {\bm\theta}_a), \quad \hat{\boldsymbol{y}} = Q(\boldsymbol{y} - \boldsymbol{\mu}) + \boldsymbol{\mu}, \quad \hat{\boldsymbol{x}} = g_s(\hat{\boldsymbol{y}}; {\bm\theta}_s),\nonumber
\end{equation}
where \( \boldsymbol{\mu} \) is the estimated entropy parameter. The entropy model uses a Hyper Encoder \( h_a(\cdot) \) and a Hyper Decoder \( h_s(\cdot) \) to transform \( \bm{y} \) into a hyper-prior \( \bm{z} \), which is quantized and transmitted as side information. The decoder network \( h_s \) converts \( \hat{\bm{z}} \) into gaussian distribution parameters \( (\bm{\mu}, \bm{\sigma}) \):
\begin{equation}
\boldsymbol{z} = h_a(\boldsymbol{y}; {\bm\phi}_a), \quad \hat{\boldsymbol{z}} = Q(\boldsymbol{z}), \quad \boldsymbol{\mu}, \boldsymbol{\sigma} = h_s(\hat{\boldsymbol{z}}; {\bm\phi}_s).\nonumber
\end{equation}

The model is trained to minimize the R-D loss function:
\begin{equation}
\begin{aligned}
\mathcal{L}= & \mathcal{R}(\hat{\boldsymbol{y}})+\mathcal{R}(\hat{\boldsymbol{z}})+\lambda \cdot \mathcal{D}(\boldsymbol{x}, \hat{\boldsymbol{x}}) \\
= & \mathbb{E}\left[-\log _2 p_{\hat{\boldsymbol{y}} \mid \hat{\boldsymbol{z}}}(\hat{\boldsymbol{y}} \mid \hat{\boldsymbol{z}})\right]+\mathbb{E}\left[-\log _2 p_{\widehat{\boldsymbol{z}}}(\widehat{\boldsymbol{z}})\right] \\ 
& +\lambda \cdot \mathbb{E}\|\boldsymbol{x}-\widehat{\boldsymbol{x}}\|_2^2,\nonumber
\end{aligned}
\end{equation}
where \( \mathcal{R}(\hat{\bm{y}}) \) and \( \mathcal{R}(\hat{\bm{z}}) \) are bitrates estimated by the entopy model, and \( \mathcal{D}(\bm{x}, \hat{\bm{x}}) \) is the distortion between the original and reconstructed images, with \( \lambda \) balancing bitrate and distortion.


\subsection{S2CFormer}
\label{S2CFormer}

In this paper, we focus on overcoming the previously sub-optimal RD-latency trade-off by rebalancing spatial interaction and channel aggregation. Existing studies have typically attributed the success of transformers in RD performance to attention-based spatial interactions, while research on channel aggregation modules for LIC remains limited. However, as illustrated in Fig. \ref{intro}, excessively complex spatial interaction modules significantly degrade decoding speed, whereas channel aggregation appears comparatively more efficient.

Consequently, we aim to determine whether channel aggregation might be the primary factor affecting RD performance. To this end, we propose a new paradigm called “S2CFormer,” which reorients LIC from spatial interaction towards channel aggregation by significantly simplifying spatial interaction modules. As depicted in Fig. \ref{backbone} (a), our proposed paradigm is a general architecture consisting of two primary components: Spatial Interaction and Channel Aggregation.

\subsubsection{S2C-Identity: The Lower Bound of S2CFormers}

To reevaluate the key factors in the R-D performance of transformers-based LIC models and explore the lower bound of S2CFormer’s performance, we instantiate the spatial interaction with the simplest operator, Identity Mapping, as illustrated in Fig. \ref{backbone} (b):
\begin{equation}
\operatorname{ Identity }({\bm X})={\bm X},\nonumber
\end{equation}
which is an identity spatial transformation, and does not perform special spatial feature interaction. We use this initialization, termed S2C-Identity, to demonstrate the effectiveness and dominance of channel aggregation.

\subsubsection{Simplified Spatial Interaction}\label{SSI}


In addition to S2C-Identity, we explore more S2CFormer instances with other spatial operators, as shown in Fig. \ref{backbone} (c-d). Instead of designing novel spatial interaction modules, we assess the model's potential for achieving SOTA performance with simplified, commonly used spatial operators. We implement two spatial interaction modules. The first is separable convolution, as proposed in \cite{yu2022metaformer, chollet2017xception, sandler2018mobilenetv2}:
\begin{equation}
\operatorname{ SepConv }( {\bm X})=\operatorname{Conv}_{\mathrm{pw}}\left(\operatorname{Conv}_{\mathrm{dw}}\left(\sigma\left(\operatorname{Conv}_{\mathrm{pw}}({\bm X})\right)\right)\right),\nonumber
\end{equation}
where $\operatorname{Conv}_{\mathrm{pw}}$ and $\operatorname{Conv}_{\mathrm{dw}}$ denote point-wise and depth-wise convolution, respectively, and $\sigma$ represents nonlinear activation function. Separable convolutions help to reduce computational cost and parameter count. This instantiation is termed as S2C-Conv.

Besides convolution, another commonly used spatial interaction module for LIC is self-attention. We employ a simplified one, vanilla window attention, to achieve basic spatial interactions. It is defined as:
\begin{equation}
\operatorname{Attention}({\bm Q}, {\bm K}, {\bm V})=\operatorname{Softmax}\left(\frac{{\bm Q} {\bm K}^{\top}}{\sqrt{d_h}}\right) {\bm V},\nonumber
\end{equation}
where ${\bm Q}, {\bm K}, {\bm V}$ represent the query, key, and value, respectively, and $d_h$ denotes the number of attention heads. It computes attention in a windowed manner, eliminating special window partitions or time-consuming window shifting operations. This instantiation is identified as S2C-Attention.

Following spatial interaction, we apply channel aggregation with Layer Normalization and a feed-forward network (FFN) consisting of linear layers, as detailed in Sec. \ref{FFN}.

\subsubsection{Advanced Channel Aggregation}\label{ACA}
\label{FFN}

\begin{figure}[ht]
\centering
\includegraphics[width=0.43\textwidth]{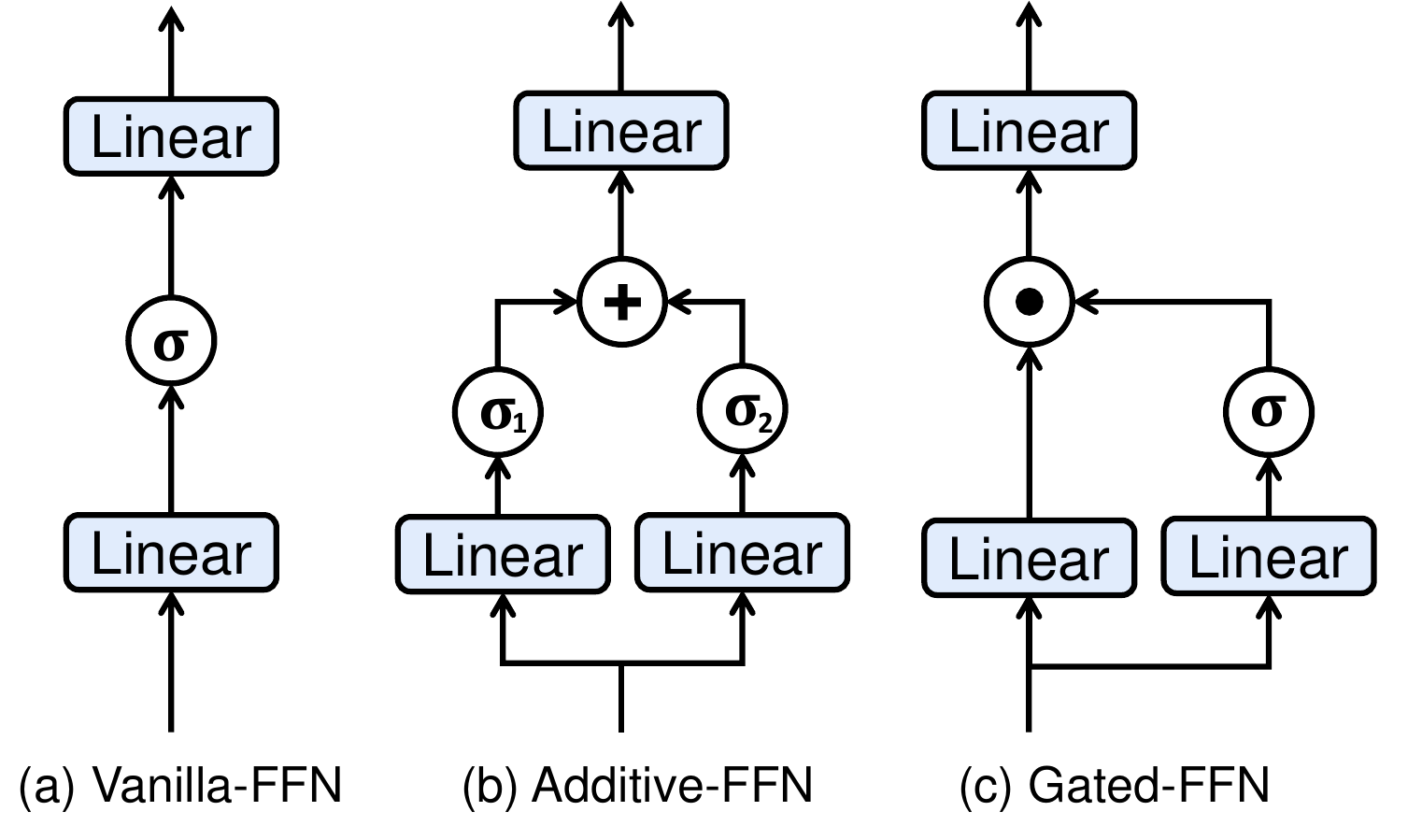}
\captionsetup{skip=1pt}  
\caption{Vanilla FFN (a) and Advanced FFNs (b-c)}
\label{mlps}
\vspace{-5pt}  
\end{figure}

This paper does not introduce complex spatial operations. Instead, we explore advanced FFN structures for channel aggregation to assess their potential benefits, aiming to enhance LIC models with minimal added complexity and achieve a better balance between performance and coding speed.
Channel aggregation is primarily achieved using MLP-based FFNs. As illustrated in Fig. \ref{mlps} (a), a vanilla FFN can be formulated as follows:
\begin{equation}
\operatorname{Vanilla\text{-}FFN}({\bm X})=\sigma\left({\bm X}  {\bm W}_{\rm in}\right)  {\bm W}_{\rm out},\nonumber
\end{equation}
where ${\bm W}_{\rm in} \in \mathbb{R}^{C \times rC}$ and ${\bm W}_{\rm out} \in \mathbb{R}^{rC \times C}$ are learnable parameters with an expansion ratio $r$ and the input channel number $C$. $\sigma$ refers to the activation function.

Inspired by previous work in NLP and low-level tasks \cite{chen2022simple, zamir2022restormer, dauphin2017language, li2023neural, li2022hybrid}, we propose two efficient yet powerful FFN structures for channel aggregation: Additive-FFN and Gated-FFN, shown in Fig. \ref{mlps} (b) and (c).

The Additive-FFN combines outputs from two activation functions via addition, applying separate nonlinear transformations to the same input to leverage the strength of multiple nonlinearities: 
\begin{equation}
\operatorname{Additive\text{-}FFN}({\bm X}) =  \left[ \sigma_1\left(  {\bm X} {\bm W}_{1}\right) + \sigma_2\left(  {\bm X} {\bm W}_{2}\right) \right]  {\bm W}_{\text{out}},\nonumber
\end{equation}
where ${\bm W}_1$ and ${\bm W}_2$ are learnable parameters, and $\sigma_1$, $\sigma_2$ are distinct activation functions.

In contrast, the Gated-FFN enhances feature interactions via the Hadamard product, performing element-wise multiplication of two linear transformation results, with one passed through an activation function:
\begin{equation}
\operatorname{Gated\text{-}FFN}({\bm X}) =  \left[ \sigma\left(  {\bm X} {\bm W}_1 \right) \odot \left(  {\bm X} {\bm W}_2\right) \right] {\bm W}_{\text{out}},\nonumber
\end{equation}
where $\odot$ denotes element-wise multiplication. For all structures, the expansion ratio $r$ is set to 4, and linear layers are implemented via $1\times1$ convolutions.

Advanced FFNs surpass vanilla FFNs in performance with comparable computational complexity and latency, highlighting the importance of efficient FFN designs for effective LIC models, beyond complex spatial interactions.

\subsubsection{S2C-Hybrid: A Superior Extension.}
The strong RD performance of S2C-Conv and S2C-Attention highlights the redundancy of prior complex spatial operations but raises new questions: While the LIC model combining these modules achieves comparable overall performance, do they perform equally at each stage? If not, could an optimal configuration maximize their strengths at specific stages, further improving LIC performance without sacrificing efficiency? Through experiments, we identified the best configuration under the current framework, with detailed results in Sec. \ref{arrangement}.

On such basis, we introduce the S2C-Hybrid model, an enhanced version of the S2CFormer-based LIC model. 
This hybrid model employs different S2C instances for different transform stages.  Specifically, we employ convolution in the first stage when feature resolution is high, and apply attention mechanisms in the later two stages. This strategy improves R-D performance without increasing time complexity or slowing down coding speed. S2C-Hybrid offers an even more excellent trade-off between R-D performance and decoding speed.

\begin{table*}[ht]
\centering                          
\setlength{\tabcolsep}{2pt}                                                      
\fontsize{8.7}{11.5}\selectfont
\renewcommand{\arraystretch}{1.25} 
    \begin{tabular}{l|ccccc|ccc|ccc}
    \noalign{\hrule height 1pt}
    \multicolumn{1}{c|}{} & \multicolumn{5}{c|}{Previous SOTA Methods} & \multicolumn{3}{c|}{S2C Basic Instances} & \multicolumn{3}{c}{S2C-Hybrid and Scaling-Up} \\
\cline{2-12} 
         & ELIC & MambaVC & FTIC & MLIC++ & TCM  & S2C-Identity & S2C-Conv & S2C-Attention &  Hybrid-S & Hybrid-M & Hybrid-L\\  \cline{1-12}
       BD-rate [Kodak] & -3.10 & -8.11 & -12.94 & -11.97 & -10.04 & -9.70 & -12.65 & -12.56 & -13.35 & -13.85 & -14.28 \\ 
       BD-rate [CLIC]  & -0.84 & - & -10.21 & -12.08 & -8.60 & -7.55  & -10.98 & -10.28 & -12.07 & -12.52 & -12.88 \\ 
       BD-rate [Tecnick]  & -7.41 & - & -13.89 & -15.13 & -10.42 & -10.70  & -14.48  & -14.08 & -15.21 & -16.34 & -17.20 \\ \cline{1-12}
        Params (M) & 33.29 & 47.88 & 69.78 & 116.48 & 75.89 & 64.63 & 66.60 & 68.42 & 68.00 & 72.73 & 79.83 \\
      FLOPs (T) & 1.74 & 2.10 & 2.38 & 2.64 & 3.74 & 2.80 & 3.13 & 3.42 & 3.20 & 3.59 & 4.17\\  
       Decoding Latency (s) & 0.335 & 0.425 & $\textgreater$10 & 0.547 & 0.542 & 0.346 & 0.356 & 0.360 & 0.357 & 0.358 & 0.362 \\ 
       Throughput(samples/s)  & 72.32 & 6.55 & 23.25 & 27.42 & 17.80 & 50.02 & 35.49 & 26.21 & 30.21 & 28.07 & 22.63 \\ 
    \noalign{\hrule height 1pt}
    \end{tabular}
    \caption{\textbf{Comprehensive comparisons of LIC models.} 
    \textbf{(a) BD-rate.} BD-rate of LIC models on different datasets, relative to VTM-21.0 \cite{JVET-AF2002}. S2C-Conv, -Attention, and -Hybrid models achieve superior R-D performance.  
    \textbf{(b) Model Params.} S2C-based models have fewer parameters compared to previous SOTA methods.
    \textbf{(c) Efficiency.} S2C-based models achieve over 30\% faster decoding speed and significantly higher training throughput. However, they are not superior in terms of FLOPs, and we will explain this issue in \ref{flops}.
    }
\vspace{-5pt}  
    \label{tab:bdrates}
\end{table*}
\section{Experiments}
\label{sec:Experiments}

\subsection{Experimental Setup}

\subsubsection{Training Details}
We train all S2CFormer-based models on the Flickr2W dataset \cite{liu2020unified} for 2 million steps with a batch size of 8, using Adam \cite{kingma2014adam} optimizer and an initial learning rate of 0.0001. 
For MSE-optimized models, we use Lagrangian multipliers \{0.0017, 0.0025, 0.0035, 0.0067, 0.0130, 0.0250, 0.050\}; for MS-SSIM-optimized models, we use \{3, 5, 8, 16, 36, 64\}. 
In the attention modules, the window size is set to 8, and depthwise convolution kernels are all of size 5. 
We set \{192, 192, 192, 320\} for channel numbers $C_1, C_2, C_3, C_4$.
 To enhance context mining and parameter aggregation within the entropy model of ELIC \cite{he2022elic}, we incorporate the S2C-Conv block and the S2C-Identity block, respectively. Unless otherwise specified, we use the Gated-FFN as a channel aggregation module for all S2CFormer-based models. All experiments are conducted on NVIDIA A100 GPUs.

\subsubsection{Evaluation}

We test our models on three datasets: Kodak image set \cite{kodak1993} with the image size of 768 × 512, Tecnick test set \cite{asuni2014testimages} with the image size of 1200 × 1200, and CLIC professional validation dataset \cite{clic2021} with 2k resolution. We take PSNR, MS-SSIM, and bits per pixel (bpp) as metrics. We utilize the BD-rate \cite{bjontegaard2001} results to quantify the average bitrate savings.


We compare our models with VTM-21.0 \cite{JVET-AF2002, bross2021overview} and several leading LIC models, including ELIC \cite{he2022elic}, MambaVC \cite{qin2024mambavc}, TCM \cite{liu2023learned}, MLIC++ \cite{jiang2023mlic} and FTIC \cite{li2023frequency}. The detailed results are presented in Table \ref{tab:bdrates}.

\subsection{Basic S2CFormer Instances}
\begin{figure}[t]
	\centering
	\begin{minipage}{0.49\linewidth}
		\centering
		\includegraphics[width=1\linewidth]{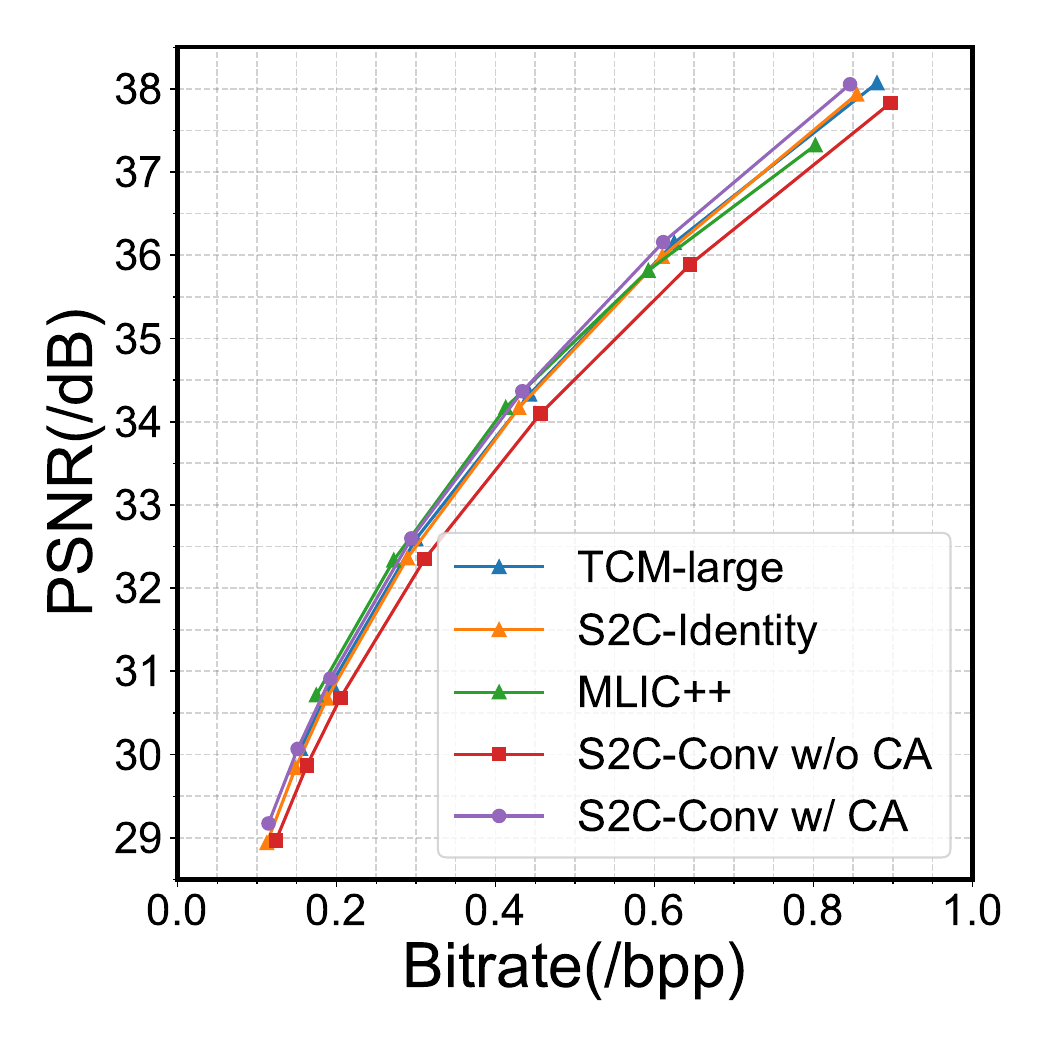}
\captionsetup{skip=0.5pt} 
		\caption*{(a)}
		\label{abl dwc} 
	\end{minipage}
	\begin{minipage}{0.49\linewidth}
		\centering
		\includegraphics[width=1\linewidth]{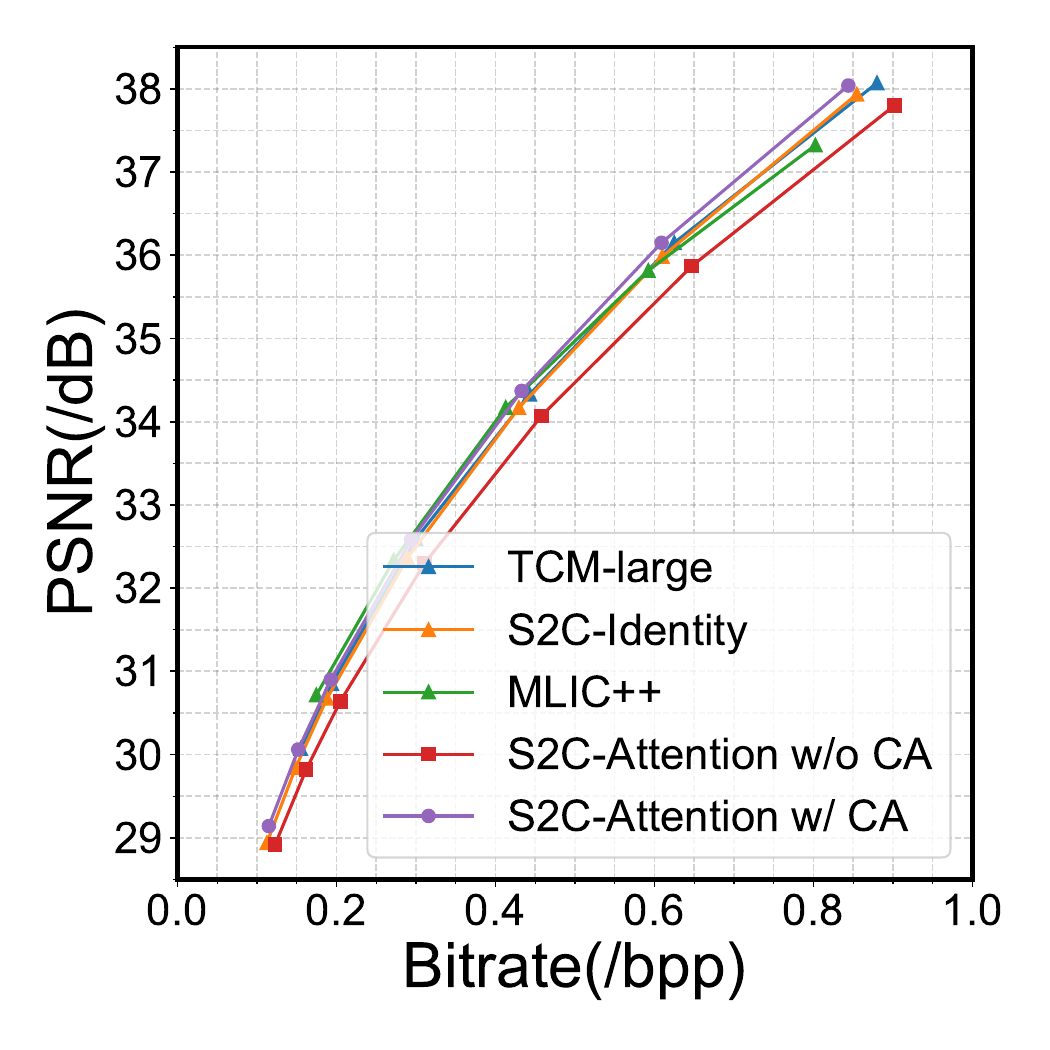}
\captionsetup{skip=0.5pt} 
		\caption*{(b)}
		\label{abl attention} 
	\end{minipage}
\captionsetup{skip=1pt} 
 \caption{\textbf{Ablation Study for Channel Aggregation.} Experiments on Kodak dataset. ``w/ FFN'' refers to S2CFormer-based models with Channel Aggregation. ``w/o CA'' represents removing all FFN modules for Channel Aggregation.}
 \label{abl ffn}
 \vspace{-10pt} 
\end{figure}

\noindent \textbf{The Lower Bound of S2CFormers.~}
S2C-Identity, which relies mainly on MLPs without spatial interactions, holds the lower performance bound for S2CFormer-based models. Surprisingly, it outperforms VTM-21.0 by 9.70\%, 10.70\%, and 7.55\% in BD-rate on Kodak, Tecnick, and CLIC datasets, respectively. It surpasses MambaVC, and nearly matches the performance of TCM-large, as shown in Fig. \ref{abl ffn}. This promising result suggests that MLP-based channel aggregation secures a solid baseline performance for LIC.

\noindent \textbf{S2CFormers With Simplified Spatial Interaction.~}
We compare the S2C-Conv and S2C-Attention models with existing models and find that both models outperform nearly all the SOTA methods. As shown in Tabel \ref{tab:bdrates}, in terms of BD-rates, both models surpass FTIC, TCM-large, and MambaVC. They also outperform MLIC++ on the Kodak dataset. Although MLIC++ performs well at lower bit rates, it underperforms at higher bitrates, showing a noticeable gap compared to our models, as shown in Fig. \ref{abl ffn}. This discrepancy arises from its focus on improving the entropy model, which benefits performance at lower bitrates. However, at higher bit rates, transform networks become more critical for capturing and reconstructing high-frequency components.

The performances of our models suggest that complex spatial interaction design may be redundant and can be simplified. S2C-Conv and S2C-Attention show near-identical performance, indicating a specific choice of spatial interaction is not that critical for LIC. Instead, FFN-based channel aggregation is the key factor that secures R-D performance.

\begin{figure}[t]
\centering
\includegraphics[width=0.43\textwidth]{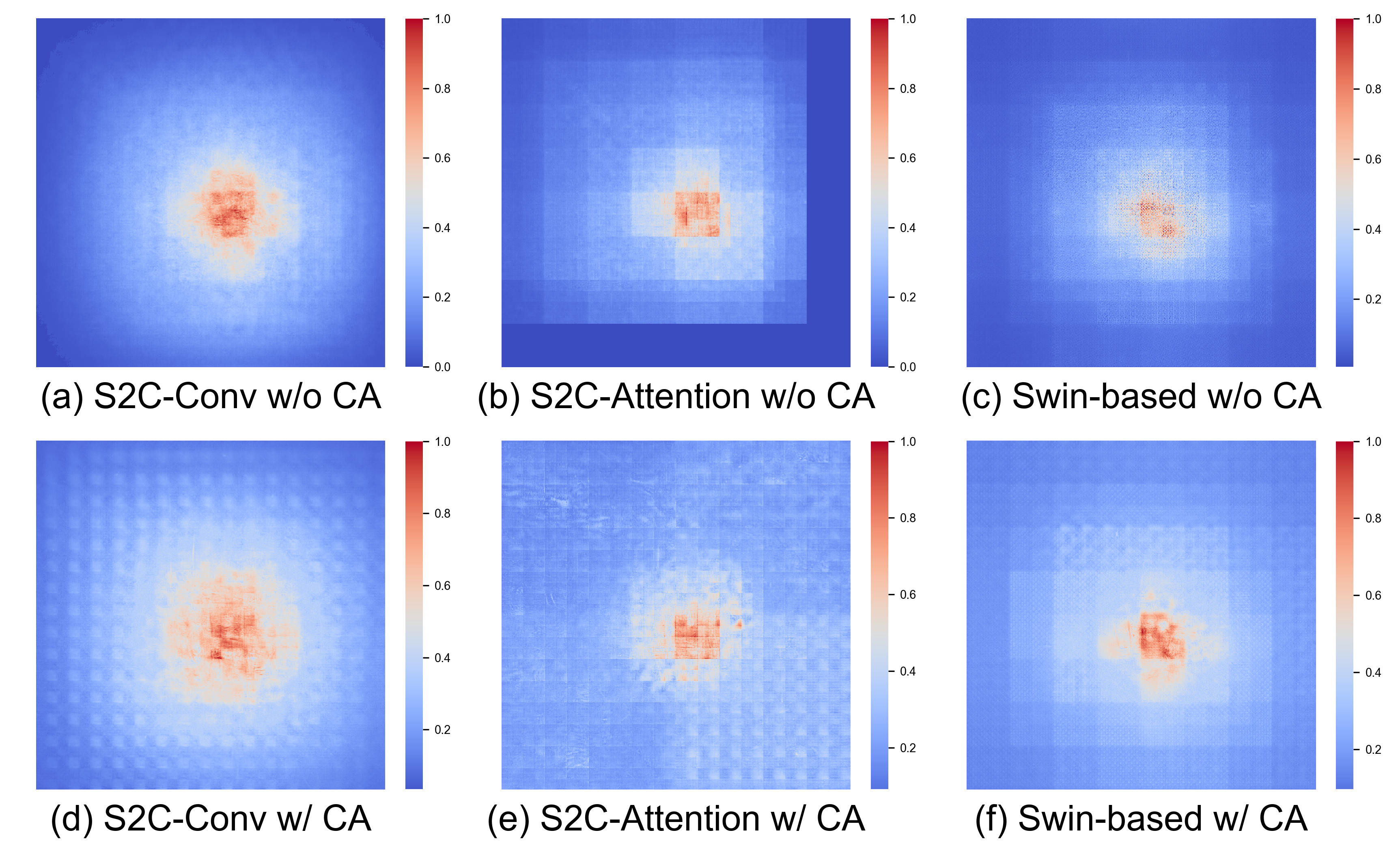}
\captionsetup{skip=1pt}  
\caption{The effective receptive fields (ERF) \cite{luo2016understanding} calculated by different models. $\text{``CA''}$ refers to Channel Aggregation module.}
\label{erf}
\vspace{-10pt}  
\end{figure}

\begin{figure*}[htbp]
    \centering
    \begin{minipage}{0.49\linewidth}
        \centering
        \includegraphics[width=1\linewidth]{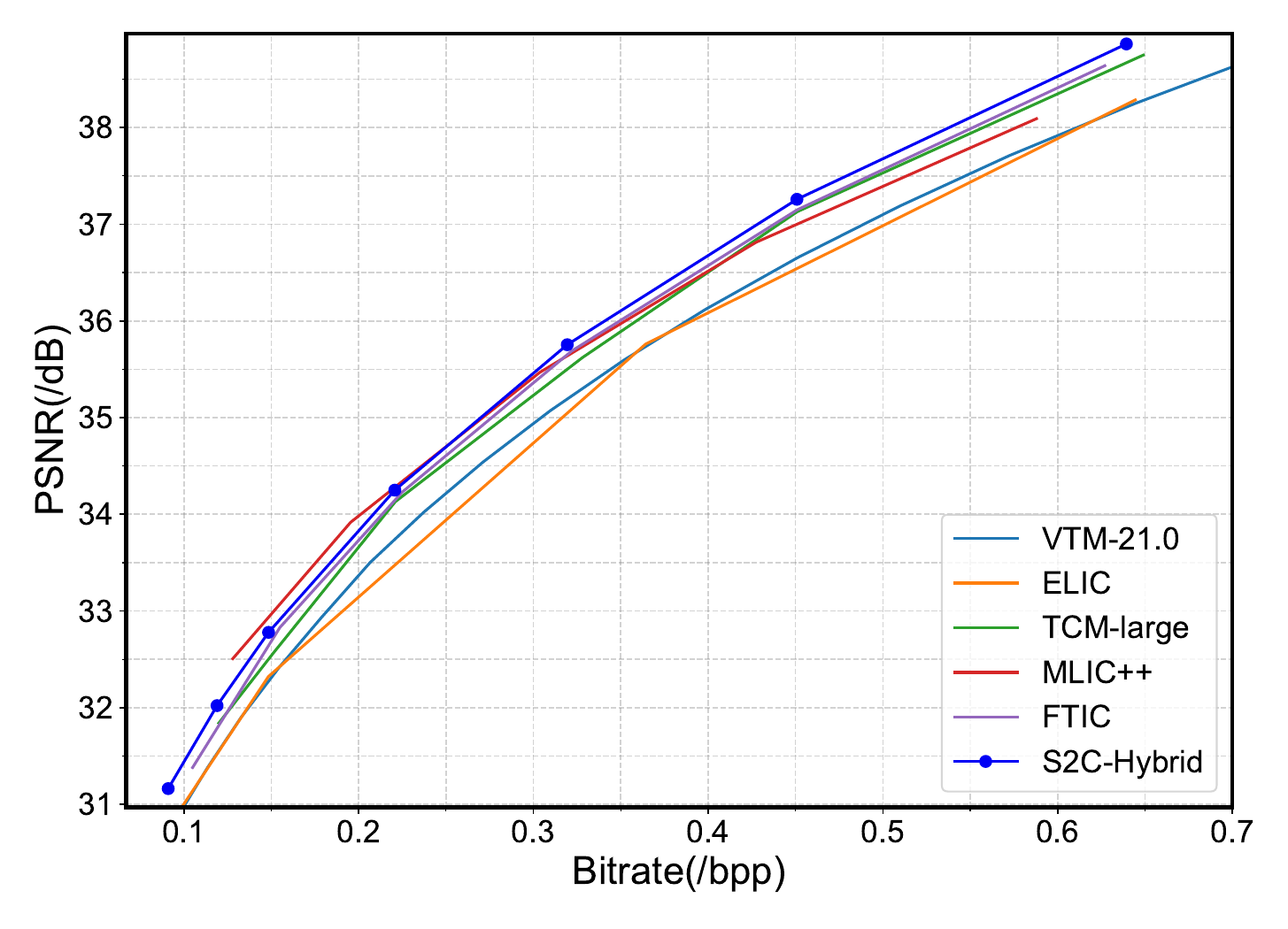}
        \captionsetup{skip=0.5pt}  
        \label{fig:clic_psnr}
    \end{minipage}
    \hfill
    \begin{minipage}{0.49\linewidth}
        \centering
        \includegraphics[width=1\linewidth]{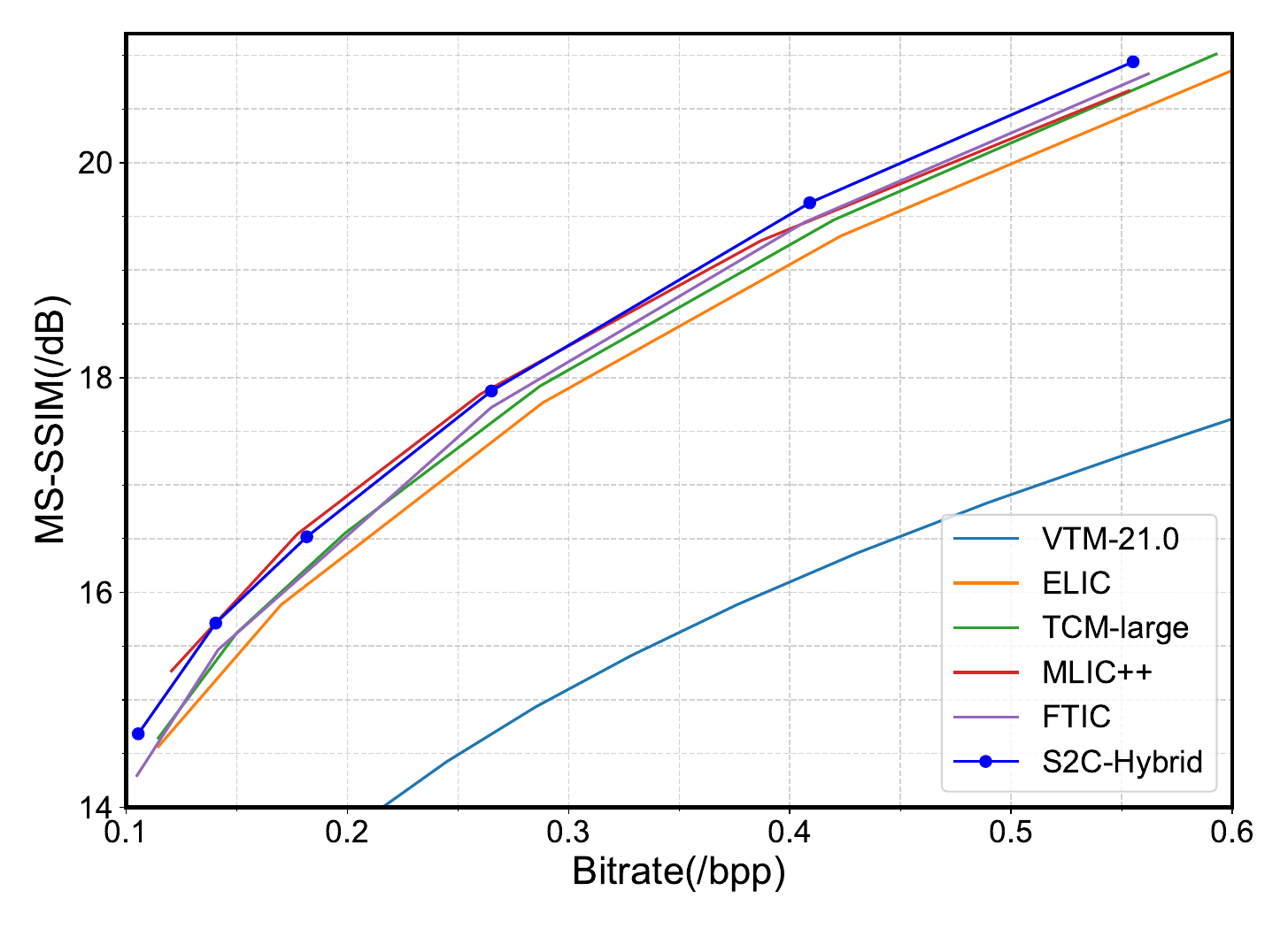}
        \captionsetup{skip=0.5pt}  
        \label{fig:clic_msssim}
    \end{minipage}
    \vspace{-10pt}  
    \captionsetup{skip=0.5pt}  
    \caption{Performance evaluation on the CLIC Professional Validation Dataset.}
    \label{rdcurves1}
    \vspace{-10pt}  
\end{figure*}
\begin{figure*}[htbp]
    \centering
    \begin{minipage}{0.49\linewidth}
        \centering
        \includegraphics[width=1\linewidth]{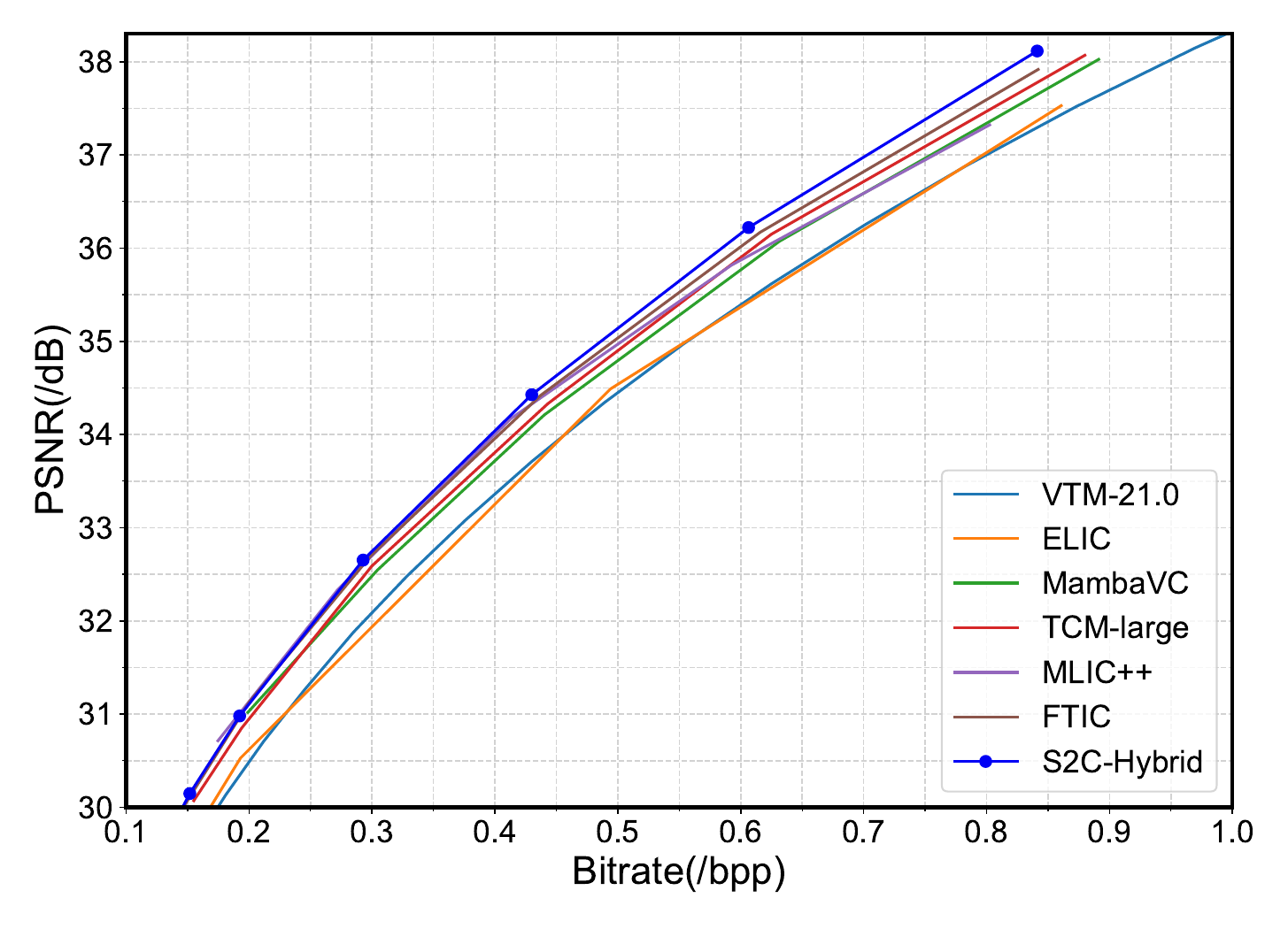}
        \captionsetup{skip=0.5pt}  
        \caption{Performance evaluation on the Kodak dataset.}
        \label{fig:kodak_psnr}
    \end{minipage}
    \hfill
    \begin{minipage}{0.49\linewidth}
        \centering
        \includegraphics[width=1\linewidth]{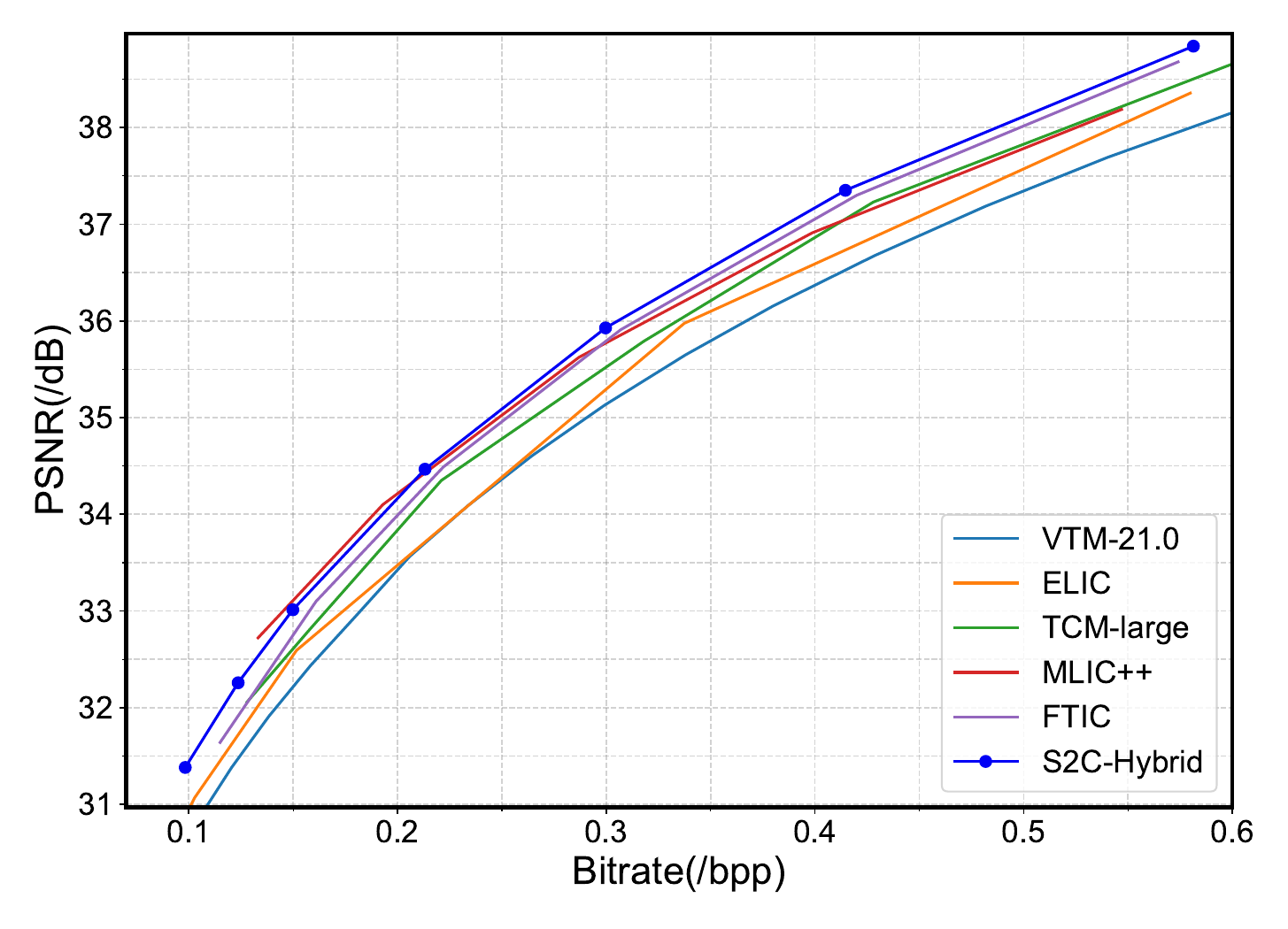}
        \captionsetup{skip=0.5pt}  
        \caption{Performance evaluation on the Tecnick dataset.}
        \label{fig:tecnick_psnr}
    \end{minipage}
    \vspace{-10pt}  
\end{figure*}

\subsection{Channel Aggregations}
\noindent \textbf{Ablations of Channel Aggregation.~} In this section, we analyze the critical role of Channel Aggregation (CA) in the S2CFormer architecture. An ablation study was conducted by removing all FFN modules from S2C-Conv and S2C-Attention. Results in Fig. \ref{abl ffn} show that FFN-enhanced separable convolution and window attention significantly outperform their counterparts, improving BD-rate by -10.57\% and -11.14\%, respectively. This highlights the substantial impact of FFNs on R-D performance. In contrast, S2C-Identity, which retains CA but eliminates spatial interaction, approaches TCM-large's performance, outperforming models without CA by -7.51\% and -8.17\% BD-rate. The study emphasizes the importance of FFNs in LIC models and supports their use for CA in future designs.

\begin{figure}[t]
	\centering
	\begin{minipage}{0.49\linewidth}
		\centering
		\includegraphics[width=1\linewidth]{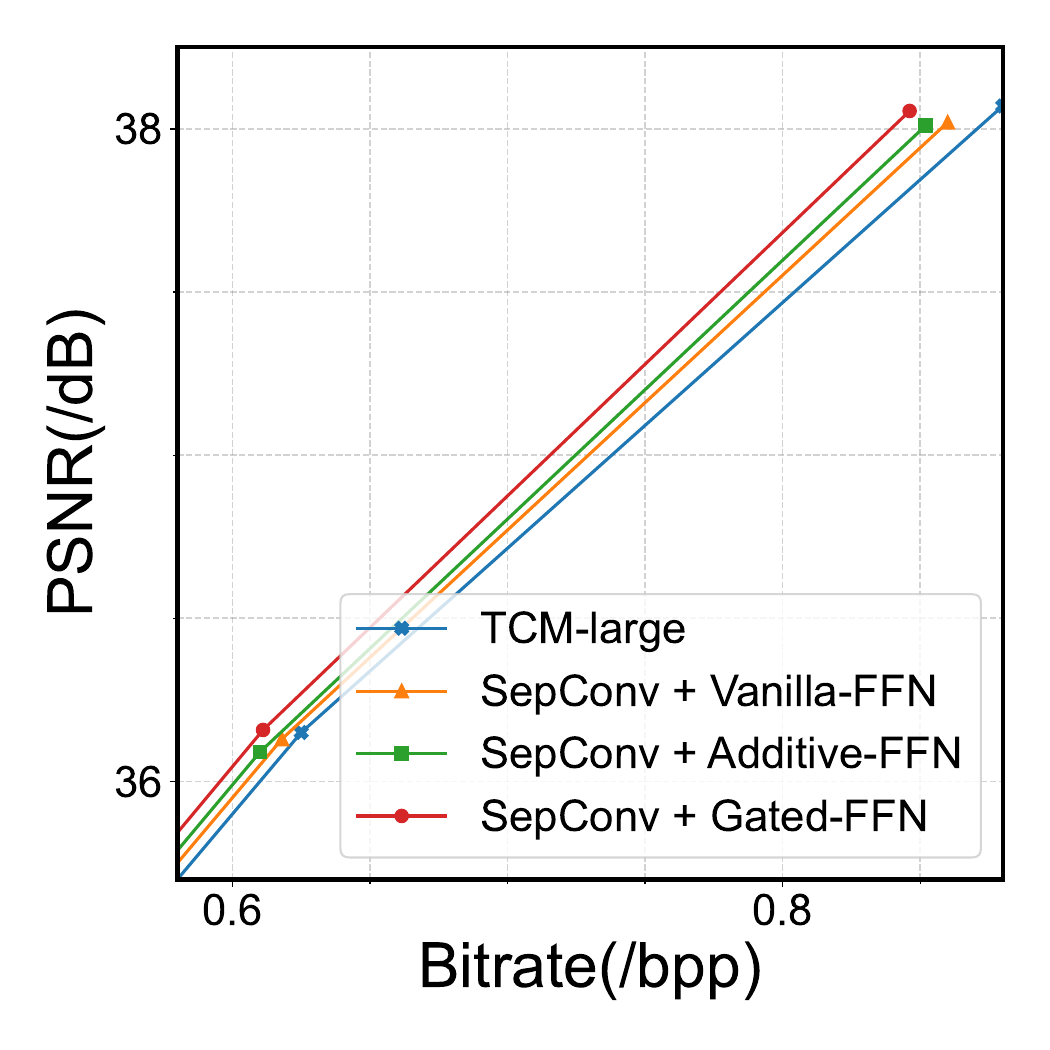}
\captionsetup{skip=0.5pt}  
		\caption*{(a)}
		\label{adv dwc}
	\end{minipage}
	\begin{minipage}{0.49\linewidth}
		\centering
		\includegraphics[width=1\linewidth]{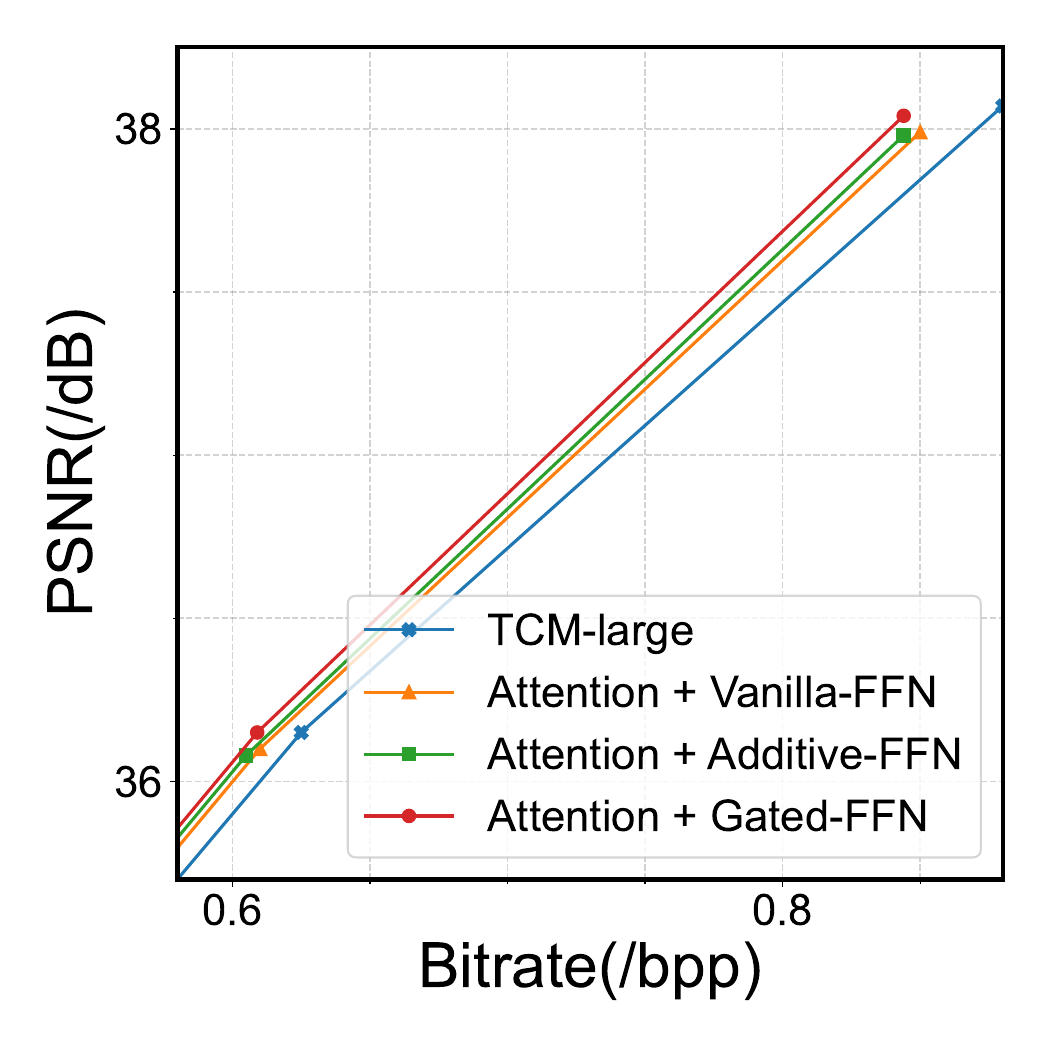}
\captionsetup{skip=0.5pt}  
		\caption*{(b)}
		\label{adv attention}
	\end{minipage}
\captionsetup{skip=1pt}  
 \caption{\textbf{Comparison of Different FFNs}. Experiments on the Kodak dataset. We provide R-D curves for S2C-Conv and -Attention models with various FFNs for Channel Aggregation.}
 \label{adv ffn}
 \vspace{-15pt}  
\end{figure}

\begin{figure*}[th]
\centering
\includegraphics[width=1\textwidth]{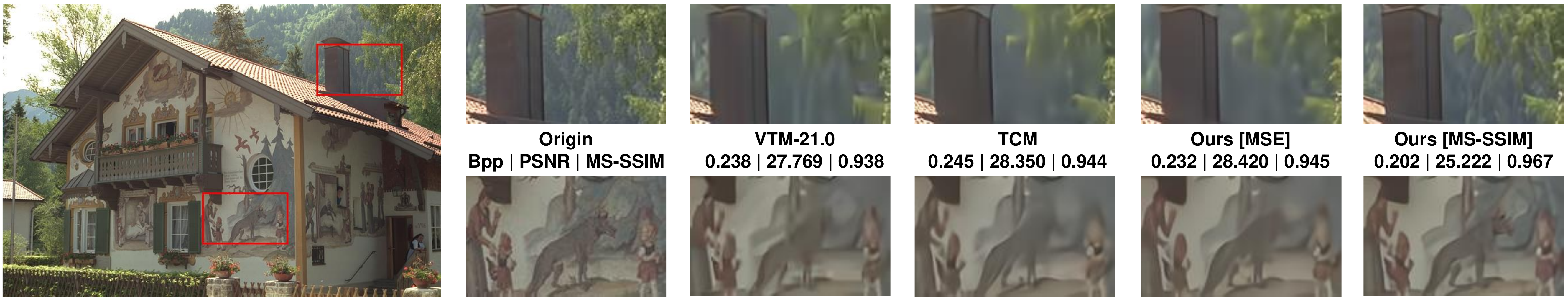}
\captionsetup{skip=1pt}  
\caption{\textbf{Visual Comparison.} This figure presents visualizations of decompressed images from the Kodak dataset using various methods. Each subfigure is labeled with ``Method $|$ Bpp $|$ PSNR $|$ MS-SSIM".}
 \vspace{-5pt}  
\label{visual1}
\end{figure*}


We analyze the Effective Receptive Field (ERF) \cite{luo2016understanding}  of networks to highlight the importance of channel aggregation. Contrary to the belief that attention mechanisms expand the receptive field and aggregate non-local information, our findings show that both S2C-Conv and S2C-Attention models exhibit smaller ERFs without channel aggregation. As shown in Fig. \ref{erf}, introducing channel aggregation significantly expands the ERF, covering more input image areas. Similarly, Swin Transformer-based LIC models rely on channel aggregation to broaden ERF, as window shifting alone does not significantly enlarge it. These results indicate that channel aggregation is crucial for activating more pixels and enhancing R-D performance, suggesting that certain complex spatial operations may be redundant.

\noindent \textbf{Effectiveness of Advanced Channel Aggregation.~}
\label{advanced} As discussed in Sec. \ref{FFN}, the structure of FFN warrants further exploration. We conduct comparison experiments on three proposed FFN structures—Vanilla-FFN, Additive-FFN, and Gated-FFN—within both the S2C-Conv and S2C-Attention models, as shown in Fig. \ref{adv ffn}. For S2C-Conv, models utilizing Vanilla-FFN, Additive-FFN, and Gated-FFN outperform VTM-21.0 by -10.88\%, -11.87\%, and -12.65\%, respectively. When window attention is applied for spatial interaction, these models achieved BD-rates of -11.27\%, -12.05\%, and -12.56\% over VTM-21.0. The decoding latency for these models is nearly identical. Gated-FFN consistently outperforms Additive-FFN, which, in turn, surpasses Vanilla-FFN. These results demonstrate that the FFN structure remains a promising area for further investigation.

\subsection{S2C-Hybrid}

\noindent \textbf{Arrangement Ablations.} \label{arrangement}
Our experiment results in Tab. \ref{tab:hybrid} show that the optimal configuration for the three-stage main transform networks is to use S2C-Conv (denoted as “C”) in the first stage and S2C-Attention (denoted as “A”) in the subsequent stages. The “[C, A, A]” configuration performs best, likely because convolution layers excel at capturing local details at high resolutions, while attention mechanisms are more effective at extracting global information as resolution decreases. This highlights that while S2C-Conv and S2C-Attention exhibit similar rate-distortion performance, their distinct capabilities at different stages significantly impact overall performance when optimally arranged.

\vspace{-5pt}
\begin{table}[ht]
\centering                          
\setlength{\tabcolsep}{4.3pt}

\fontsize{8.35}{10}\selectfont
\renewcommand{\arraystretch}{1} 
    \begin{tabular}{c|cc|ccc}
    \hline
      Stages[1,2,3]   & [C,C,C] & [A,A,A] & [A,C,C] & [C,C,A] & [C,A,A]\\  \cline{1-6}
       BD-rate (\%) & -12.65 & -12.56 & -12.32 & {-12.87} & \textbf{-13.35} \\  
    \hline
    \end{tabular}
\vspace{-5pt}
    \caption{S2C-Hybrid Arrangement Ablations 
    }
\vspace{-20pt} 
    \label{tab:hybrid}
\end{table}


\noindent\textbf{Performance and Scaling-Up.} 
As shown in Table \ref{tab:bdrates}, the standard S2C-Hybrid model variant, Hybrid-S, which uses a configuration of \{3, 3, 3\} for stages $L_1$, $L_2$, and $L_3$, has achieved state-of-the-art performance across all three datasets while maintaining low decoding latency. 
Subsequently, we scaled up the S2C-Hybrid model to further explore the trade-off between model complexity and rate-distortion performance.  Hybrid-M uses \{3, 5, 5\} blocks and Hybrid-L uses \{3, 8, 8\} blocks for stages \(L_1\), \(L_2\), and \(L_3\). As reported in Table \ref{tab:bdrates}, stacking more blocks in the last two stages consistently improves performance.
The RD curves for Hybrid-L are presented in Figures \ref{rdcurves1}–\ref{fig:tecnick_psnr}.

\noindent\textbf{Visual Comparison} As illustrated in Fig. \ref{visual1}, comparisons with VTM-21.0 and TCM demonstrate that our proposed method achieves superior detail preservation.


\subsection{Model efficiency}
\label{efficiency}

We calculate the FLOPs and decoding latency for various models using 2K images, and evaluate training throughput with a batch size of 8 and patch size of 256$\times$256. Training throughput is directly correlated to the training time of LIC models. Numeric results are provided in Table \ref{tab:bdrates}.

\noindent \textbf{Decoding Latency.~} S2CFormer-based models demonstrate significant advantages in decoding latency. All of our models are quite fast, with even the slowest one, S2C-Hybrid, achieving a decoding speed that is more than 30\% faster than MLIC++ and TCM-large.

\noindent \textbf{Training Throughput.~} S2CFormer-based models also significantly increase training throughput, S2C-Identity is competitive with TCM-large in R-D performance, while improving training speed by 180\%, which cuts the total training time from 10.4 days to 3.7 days. Similarly, S2C-Conv outperforms FTIC in R-D performance, and accelerates training by 52\%, saving over 65 hours of training time.

\noindent \textbf{FLOPs.~} \label{flops}
For VAE-based LIC models, the number of channels in the first stage significantly influences the FLOPs. In our work, we set all channel numbers to 192, following the classical setting of LIC models \cite{balle2016end, minnen2018joint, balle2018variational, cheng2020learned, he2022elic}, since optimizing channel settings falls beyond the scope of this study. Although some leading methods reduce the first-stage channel count to 128 or fewer \cite{qin2024mambavc, li2023frequency} to lower FLOPs, these models remain considerably slower than our S2CFormer-based models in both training and decoding.

This paper focuses on rebalancing spatial interactions and channel aggregation for low decoding latency. In this context, S2CFormer serves as a paradigm that permits flexible channel adjustments to reduce FLOPs. To demonstrate this, we propose a variant named Hybrid-T, designed for low-FLOP operation. Specifically, we set the channel numbers $C_1$, $C_2$, and $C_3$ to 96, 192, and 256, and the block numbers $L_1$, $L_2$, and $L_3$ to 3, 5, and 8, respectively. Hybrid-T achieves a SOTA BD-rate of -13.32\% on the Kodak dataset, while its FLOPs are only 2.63T—a value relatively low compared to previous SOTA methods. These results indicate that our S2CFormer is not a fixed design, and we can achieve even more efficient S2C models by adjusting the channel numbers to find the optimal configuration in the future.

\section{Conclusion}
\label{sec:Conclusion}

This paper tackles the trade-offs between decoding latency and rate-distortion performance in transformer-based LIC models. We emphasize the critical role of channel aggregation and demonstrate through extensive experiments that channel operations—rather than some time-consuming spatial interactions—are key to achieving high-performance LIC models. Based on these insights, we propose the S2CFormer paradigm, which streamlines spatial operations while substantially enhancing channel aggregation.

Building on these findings, we evaluate several S2CFormer instances. Experimental results from S2C-Conv and S2C-Attention confirm that combining simpler spatial interactions with effective channel aggregation not only delivers state-of-the-art compression performance but also significantly accelerates decoding speed. Furthermore, the advanced S2C-Hybrid model, which integrates the strengths of multiple S2C variants, establishes a new benchmark for LIC by outperforming existing models on standard datasets such as Kodak, Tecnick, and CLIC Professional Validation. Overall, this work sets a new standard for efficient, high-performance LIC and lays a robust foundation for future research on optimized channel aggregation strategies.


{
    \small
    \bibliographystyle{ieeenat_fullname}
    \bibliography{main}

\begin{thebibliography}{58}
\providecommand{\natexlab}[1]{#1}
\providecommand{\url}[1]{\texttt{#1}}
\expandafter\ifx\csname urlstyle\endcsname\relax
  \providecommand{\doi}[1]{doi: #1}\else
  \providecommand{\doi}{doi: \begingroup \urlstyle{rm}\Url}\fi

\bibitem[Asuni et~al.(2014)Asuni, Giachetti, et~al.]{asuni2014testimages}
Nicola Asuni, Andrea Giachetti, et~al.
\newblock Testimages: a large-scale archive for testing visual devices and basic image processing algorithms.
\newblock In \emph{STAG}, pages 63--70, 2014.

\bibitem[Ball{\'e} et~al.(2016)Ball{\'e}, Laparra, and Simoncelli]{balle2016end}
Johannes Ball{\'e}, Valero Laparra, and Eero~P Simoncelli.
\newblock End-to-end optimized image compression.
\newblock \emph{arXiv preprint arXiv:1611.01704}, 2016.

\bibitem[Ball{\'e} et~al.(2018)Ball{\'e}, Minnen, Singh, Hwang, and Johnston]{balle2018variational}
Johannes Ball{\'e}, David Minnen, Saurabh Singh, Sung~Jin Hwang, and Nick Johnston.
\newblock Variational image compression with a scale hyperprior.
\newblock \emph{arXiv preprint arXiv:1802.01436}, 2018.

\bibitem[B{\'e}gaint et~al.(2020)B{\'e}gaint, Racap{\'e}, Feltman, and Pushparaja]{begaint2020compressai}
Jean B{\'e}gaint, Fabien Racap{\'e}, Simon Feltman, and Akshay Pushparaja.
\newblock Compressai: a pytorch library and evaluation platform for end-to-end compression research.
\newblock \emph{arXiv e-prints}, pages arXiv--2011, 2020.

\bibitem[Bjontegaard(2001)]{bjontegaard2001}
Gisle Bjontegaard.
\newblock Calculation of average psnr differences between rd-curves.
\newblock In \emph{VCEG-M33}, 2001.

\bibitem[Bross et~al.(2021)Bross, Wang, Ye, Liu, Chen, Sullivan, and Ohm]{bross2021overview}
Benjamin Bross, Ye-Kui Wang, Yan Ye, Shan Liu, Jianle Chen, Gary~J Sullivan, and Jens-Rainer Ohm.
\newblock Overview of the versatile video coding (vvc) standard and its applications.
\newblock \emph{IEEE Transactions on Circuits and Systems for Video Technology}, 31\penalty0 (10):\penalty0 3736--3764, 2021.

\bibitem[Browne et~al.()Browne, Ye, and Kim]{JVET-AF2002}
A. Browne, Y. Ye, and S. Kim.
\newblock Algorithm description for versatile video coding and test model 21 (vtm 21), document jvet-af2002.
\newblock In \emph{Joint Video Experts Team (JVET) of ITU-T SG 16 WP 3 and ISO/IEC JTC 1/SC 29/WG 11, 32nd Meeting}, Hannover.

\bibitem[Chen et~al.(2022{\natexlab{a}})Chen, Xu, and Wang]{chen2022two}
Fangdong Chen, Yumeng Xu, and Li Wang.
\newblock Two-stage octave residual network for end-to-end image compression.
\newblock In \emph{Proceedings of the AAAI Conference on Artificial Intelligence}, pages 3922--3929, 2022{\natexlab{a}}.

\bibitem[Chen et~al.(2022{\natexlab{b}})Chen, Chu, Zhang, and Sun]{chen2022simple}
Liangyu Chen, Xiaojie Chu, Xiangyu Zhang, and Jian Sun.
\newblock Simple baselines for image restoration.
\newblock In \emph{European conference on computer vision}, pages 17--33. Springer, 2022{\natexlab{b}}.

\bibitem[Cheng et~al.(2020)Cheng, Sun, Takeuchi, and Katto]{cheng2020learned}
Zhengxue Cheng, Heming Sun, Masaru Takeuchi, and Jiro Katto.
\newblock Learned image compression with discretized gaussian mixture likelihoods and attention modules.
\newblock In \emph{Proceedings of the IEEE/CVF conference on computer vision and pattern recognition}, pages 7939--7948, 2020.

\bibitem[Chollet(2017)]{chollet2017xception}
Fran{\c{c}}ois Chollet.
\newblock Xception: Deep learning with depthwise separable convolutions.
\newblock In \emph{Proceedings of the IEEE conference on computer vision and pattern recognition}, pages 1251--1258, 2017.

\bibitem[CLIC(2021)]{clic2021}
CLIC.
\newblock Workshop and challenge on learned image compression.
\newblock In \emph{Proceedings of the IEEE/CVF Conference on Computer Vision and Pattern Recognition}, 2021.

\bibitem[Dauphin et~al.(2017)Dauphin, Fan, Auli, and Grangier]{dauphin2017language}
Yann~N Dauphin, Angela Fan, Michael Auli, and David Grangier.
\newblock Language modeling with gated convolutional networks.
\newblock In \emph{International conference on machine learning}, pages 933--941. PMLR, 2017.

\bibitem[Dong et~al.(2022)Dong, Bao, Chen, Zhang, Yu, Yuan, Chen, and Guo]{dong2022cswin}
Xiaoyi Dong, Jianmin Bao, Dongdong Chen, Weiming Zhang, Nenghai Yu, Lu Yuan, Dong Chen, and Baining Guo.
\newblock Cswin transformer: A general vision transformer backbone with cross-shaped windows.
\newblock In \emph{Proceedings of the IEEE/CVF conference on computer vision and pattern recognition}, pages 12124--12134, 2022.

\bibitem[Dosovitskiy(2020)]{dosovitskiy2020image}
Alexey Dosovitskiy.
\newblock An image is worth 16x16 words: Transformers for image recognition at scale.
\newblock \emph{arXiv preprint arXiv:2010.11929}, 2020.

\bibitem[Fu et~al.(2023)Fu, Liang, Liang, Li, Zhang, and Han]{fu2023asymmetric}
Haisheng Fu, Feng Liang, Jie Liang, Binglin Li, Guohe Zhang, and Jingning Han.
\newblock Asymmetric learned image compression with multi-scale residual block, importance scaling, and post-quantization filtering.
\newblock \emph{IEEE Transactions on Circuits and Systems for Video Technology}, 33\penalty0 (8):\penalty0 4309--4321, 2023.

\bibitem[Gao et~al.(2021)Gao, You, Pan, Han, Zhang, Dai, and Lee]{gao2021neural}
Ge Gao, Pei You, Rong Pan, Shunyuan Han, Yuanyuan Zhang, Yuchao Dai, and Hojae Lee.
\newblock Neural image compression via attentional multi-scale back projection and frequency decomposition.
\newblock In \emph{Proceedings of the IEEE/CVF International Conference on Computer Vision}, pages 14677--14686, 2021.

\bibitem[Han et~al.(2024)Han, Jiang, Li, Deng, Xu, Zhu, and Gu]{han2024causal}
Minghao Han, Shiyin Jiang, Shengxi Li, Xin Deng, Mai Xu, Ce Zhu, and Shuhang Gu.
\newblock Causal context adjustment loss for learned image compression.
\newblock In \emph{The Thirty-eighth Annual Conference on Neural Information Processing Systems}, 2024.

\bibitem[He et~al.(2024)He, Chen, Lu, Song, and Zhang]{he2024s4d}
Bing He, Yunuo Chen, Guo Lu, Li Song, and Wenjun Zhang.
\newblock S4d: Streaming 4d real-world reconstruction with gaussians and 3d control points.
\newblock \emph{arXiv preprint arXiv:2408.13036}, 2024.

\bibitem[He et~al.(2021)He, Zheng, Sun, Wang, and Qin]{he2021checkerboard}
Dailan He, Yaoyan Zheng, Baocheng Sun, Yan Wang, and Hongwei Qin.
\newblock Checkerboard context model for efficient learned image compression.
\newblock In \emph{Proceedings of the IEEE/CVF Conference on Computer Vision and Pattern Recognition}, pages 14771--14780, 2021.

\bibitem[He et~al.(2022)He, Yang, Peng, Ma, Qin, and Wang]{he2022elic}
Dailan He, Ziming Yang, Weikun Peng, Rui Ma, Hongwei Qin, and Yan Wang.
\newblock Elic: Efficient learned image compression with unevenly grouped space-channel contextual adaptive coding.
\newblock In \emph{Proceedings of the IEEE/CVF Conference on Computer Vision and Pattern Recognition}, pages 5718--5727, 2022.

\bibitem[Jiang and Wang(2023)]{jiang2023mlic++}
Wei Jiang and Ronggang Wang.
\newblock {MLIC}\${\textasciicircum}\{++\}\$: Linear complexity multi-reference entropy modeling for learned image compression.
\newblock In \emph{ICML 2023 Workshop Neural Compression: From Information Theory to Applications}, 2023.

\bibitem[Jiang et~al.(2023)Jiang, Yang, Zhai, Ning, Gao, and Wang]{jiang2023mlic}
Wei Jiang, Jiayu Yang, Yongqi Zhai, Peirong Ning, Feng Gao, and Ronggang Wang.
\newblock Mlic: Multi-reference entropy model for learned image compression.
\newblock In \emph{Proceedings of the 31st ACM International Conference on Multimedia}, pages 7618--7627, 2023.

\bibitem[Kingma(2014)]{kingma2014adam}
Diederik~P Kingma.
\newblock Adam: A method for stochastic optimization.
\newblock \emph{arXiv preprint arXiv:1412.6980}, 2014.

\bibitem[Kodak(1993)]{kodak1993}
Eastman Kodak.
\newblock Kodak lossless true color image suite (photocd pcd0992), 1993.
\newblock Available from http://r0k.us/graphics/kodak/.

\bibitem[Koyuncu et~al.(2022)Koyuncu, Gao, Boev, Gaikov, Alshina, and Steinbach]{koyuncu2022contextformer}
A~Burakhan Koyuncu, Han Gao, Atanas Boev, Georgii Gaikov, Elena Alshina, and Eckehard Steinbach.
\newblock Contextformer: A transformer with spatio-channel attention for context modeling in learned image compression.
\newblock In \emph{European Conference on Computer Vision}, pages 447--463. Springer, 2022.

\bibitem[Lee-Thorp et~al.(2021)Lee-Thorp, Ainslie, Eckstein, and Ontanon]{lee2021fnet}
James Lee-Thorp, Joshua Ainslie, Ilya Eckstein, and Santiago Ontanon.
\newblock Fnet: Mixing tokens with fourier transforms.
\newblock \emph{arXiv preprint arXiv:2105.03824}, 2021.

\bibitem[Li et~al.(2023{\natexlab{a}})Li, Li, Dai, Li, Zou, and Xiong]{li2023frequency}
Han Li, Shaohui Li, Wenrui Dai, Chenglin Li, Junni Zou, and Hongkai Xiong.
\newblock Frequency-aware transformer for learned image compression.
\newblock \emph{arXiv preprint arXiv:2310.16387}, 2023{\natexlab{a}}.

\bibitem[Li et~al.(2022)Li, Li, and Lu]{li2022hybrid}
Jiahao Li, Bin Li, and Yan Lu.
\newblock Hybrid spatial-temporal entropy modelling for neural video compression.
\newblock In \emph{Proceedings of the 30th ACM International Conference on Multimedia}, pages 1503--1511, 2022.

\bibitem[Li et~al.(2023{\natexlab{b}})Li, Li, and Lu]{li2023neural}
Jiahao Li, Bin Li, and Yan Lu.
\newblock Neural video compression with diverse contexts.
\newblock In \emph{Proceedings of the IEEE/CVF Conference on Computer Vision and Pattern Recognition}, pages 22616--22626, 2023{\natexlab{b}}.

\bibitem[Liang et~al.(2021)Liang, Cao, Sun, Zhang, Van~Gool, and Timofte]{liang2021swinir}
Jingyun Liang, Jiezhang Cao, Guolei Sun, Kai Zhang, Luc Van~Gool, and Radu Timofte.
\newblock Swinir: Image restoration using swin transformer.
\newblock In \emph{Proceedings of the IEEE/CVF international conference on computer vision}, pages 1833--1844, 2021.

\bibitem[Liu et~al.(2020)Liu, Lu, Hu, and Xu]{liu2020unified}
Jiaheng Liu, Guo Lu, Zhihao Hu, and Dong Xu.
\newblock A unified end-to-end framework for efficient deep image compression.
\newblock \emph{arXiv preprint arXiv:2002.03370}, 2020.

\bibitem[Liu et~al.(2023)Liu, Sun, and Katto]{liu2023learned}
Jinming Liu, Heming Sun, and Jiro Katto.
\newblock Learned image compression with mixed transformer-cnn architectures.
\newblock In \emph{Proceedings of the IEEE/CVF conference on computer vision and pattern recognition}, pages 14388--14397, 2023.

\bibitem[Liu et~al.(2021)Liu, Lin, Cao, Hu, Wei, Zhang, Lin, and Guo]{liu2021swin}
Ze Liu, Yutong Lin, Yue Cao, Han Hu, Yixuan Wei, Zheng Zhang, Stephen Lin, and Baining Guo.
\newblock Swin transformer: Hierarchical vision transformer using shifted windows.
\newblock In \emph{Proceedings of the IEEE/CVF international conference on computer vision}, pages 10012--10022, 2021.

\bibitem[Liu et~al.(2022)Liu, Hu, Lin, Yao, Xie, Wei, Ning, Cao, Zhang, Dong, et~al.]{liu2022swin}
Ze Liu, Han Hu, Yutong Lin, Zhuliang Yao, Zhenda Xie, Yixuan Wei, Jia Ning, Yue Cao, Zheng Zhang, Li Dong, et~al.
\newblock Swin transformer v2: Scaling up capacity and resolution.
\newblock In \emph{Proceedings of the IEEE/CVF conference on computer vision and pattern recognition}, pages 12009--12019, 2022.

\bibitem[Lu et~al.(2022)Lu, Guo, Shi, Cao, and Ma]{lu2022transformer}
Ming Lu, Peiyao Guo, Huiqing Shi, Chuntong Cao, and Zhan Ma.
\newblock Transformer-based image compression.
\newblock In \emph{2022 Data Compression Conference (DCC)}, pages 469--469. IEEE, 2022.

\bibitem[Luo et~al.(2016)Luo, Li, Urtasun, and Zemel]{luo2016understanding}
Wenjie Luo, Yujia Li, Raquel Urtasun, and Richard Zemel.
\newblock Understanding the effective receptive field in deep convolutional neural networks.
\newblock \emph{Advances in neural information processing systems}, 29, 2016.

\bibitem[Ma et~al.(2019)Ma, Liu, Xiong, and Wu]{ma2019iwave}
Haichuan Ma, Dong Liu, Ruiqin Xiong, and Feng Wu.
\newblock iwave: Cnn-based wavelet-like transform for image compression.
\newblock \emph{IEEE Transactions on Multimedia}, 22\penalty0 (7):\penalty0 1667--1679, 2019.

\bibitem[Ma et~al.(2020)Ma, Liu, Yan, Li, and Wu]{ma2020end}
Haichuan Ma, Dong Liu, Ning Yan, Houqiang Li, and Feng Wu.
\newblock End-to-end optimized versatile image compression with wavelet-like transform.
\newblock \emph{IEEE Transactions on Pattern Analysis and Machine Intelligence}, 44\penalty0 (3):\penalty0 1247--1263, 2020.

\bibitem[Mentzer et~al.(2022)Mentzer, Toderici, Minnen, Hwang, Caelles, Lucic, and Agustsson]{mentzer2022vct}
Fabian Mentzer, George Toderici, David Minnen, Sung-Jin Hwang, Sergi Caelles, Mario Lucic, and Eirikur Agustsson.
\newblock Vct: A video compression transformer.
\newblock \emph{arXiv preprint arXiv:2206.07307}, 2022.

\bibitem[Minnen and Singh(2020)]{minnen2020channel}
David Minnen and Saurabh Singh.
\newblock Channel-wise autoregressive entropy models for learned image compression.
\newblock In \emph{2020 IEEE International Conference on Image Processing (ICIP)}, pages 3339--3343. IEEE, 2020.

\bibitem[Minnen et~al.(2018)Minnen, Ball{\'e}, and Toderici]{minnen2018joint}
David Minnen, Johannes Ball{\'e}, and George~D Toderici.
\newblock Joint autoregressive and hierarchical priors for learned image compression.
\newblock \emph{Advances in neural information processing systems}, 31, 2018.

\bibitem[Qian et~al.(2022)Qian, Lin, Sun, Tan, and Jin]{qian2022entroformer}
Yichen Qian, Ming Lin, Xiuyu Sun, Zhiyu Tan, and Rong Jin.
\newblock Entroformer: A transformer-based entropy model for learned image compression.
\newblock \emph{arXiv preprint arXiv:2202.05492}, 2022.

\bibitem[Qin et~al.(2024)Qin, Wang, Zhou, Chen, Luo, An, Dai, Xia, and Wang]{qin2024mambavc}
Shiyu Qin, Jinpeng Wang, Yimin Zhou, Bin Chen, Tianci Luo, Baoyi An, Tao Dai, Shutao Xia, and Yaowei Wang.
\newblock Mambavc: Learned visual compression with selective state spaces.
\newblock \emph{arXiv preprint arXiv:2405.15413}, 2024.

\bibitem[Ren et~al.(2023)Ren, Deng, Chen, Lou, and Zhang]{ren2023bayesian}
Qihan Ren, Huiqi Deng, Yunuo Chen, Siyu Lou, and Quanshi Zhang.
\newblock Bayesian neural networks avoid encoding complex and perturbation-sensitive concepts.
\newblock In \emph{International Conference on Machine Learning}, pages 28889--28913. PMLR, 2023.

\bibitem[Sandler et~al.(2018)Sandler, Howard, Zhu, Zhmoginov, and Chen]{sandler2018mobilenetv2}
Mark Sandler, Andrew Howard, Menglong Zhu, Andrey Zhmoginov, and Liang-Chieh Chen.
\newblock Mobilenetv2: Inverted residuals and linear bottlenecks.
\newblock In \emph{Proceedings of the IEEE conference on computer vision and pattern recognition}, pages 4510--4520, 2018.

\bibitem[Touvron et~al.(2021)Touvron, Cord, Douze, Massa, Sablayrolles, and J{\'e}gou]{touvron2021training}
Hugo Touvron, Matthieu Cord, Matthijs Douze, Francisco Massa, Alexandre Sablayrolles, and Herv{\'e} J{\'e}gou.
\newblock Training data-efficient image transformers \& distillation through attention.
\newblock In \emph{International conference on machine learning}, pages 10347--10357. PMLR, 2021.

\bibitem[Vaswani(2017)]{vaswani2017attention}
A Vaswani.
\newblock Attention is all you need.
\newblock \emph{Advances in Neural Information Processing Systems}, 2017.

\bibitem[Xia et~al.(2022)Xia, Pan, Song, Li, and Huang]{xia2022vision}
Zhuofan Xia, Xuran Pan, Shiji Song, Li~Erran Li, and Gao Huang.
\newblock Vision transformer with deformable attention.
\newblock In \emph{Proceedings of the IEEE/CVF conference on computer vision and pattern recognition}, pages 4794--4803, 2022.

\bibitem[Xie et~al.(2021)Xie, Cheng, and Chen]{xie2021enhanced}
Yueqi Xie, Ka~Leong Cheng, and Qifeng Chen.
\newblock Enhanced invertible encoding for learned image compression.
\newblock In \emph{Proceedings of the 29th ACM international conference on multimedia}, pages 162--170, 2021.

\bibitem[Yu et~al.(2022)Yu, Luo, Zhou, Si, Zhou, Wang, Feng, and Yan]{yu2022metaformer}
Weihao Yu, Mi Luo, Pan Zhou, Chenyang Si, Yichen Zhou, Xinchao Wang, Jiashi Feng, and Shuicheng Yan.
\newblock Metaformer is actually what you need for vision.
\newblock In \emph{Proceedings of the IEEE/CVF conference on computer vision and pattern recognition}, pages 10819--10829, 2022.

\bibitem[Yu et~al.(2023)Yu, Si, Zhou, Luo, Zhou, Feng, Yan, and Wang]{yu2023metaformer}
Weihao Yu, Chenyang Si, Pan Zhou, Mi Luo, Yichen Zhou, Jiashi Feng, Shuicheng Yan, and Xinchao Wang.
\newblock Metaformer baselines for vision.
\newblock \emph{IEEE Transactions on Pattern Analysis and Machine Intelligence}, 2023.

\bibitem[Zafari et~al.(2023)Zafari, Khoshkhahtinat, Mehta, Saadabadi, Akyash, and Nasrabadi]{zafari2023frequency}
Ali Zafari, Atefeh Khoshkhahtinat, Piyush Mehta, Mohammad Saeed~Ebrahimi Saadabadi, Mohammad Akyash, and Nasser~M Nasrabadi.
\newblock Frequency disentangled features in neural image compression.
\newblock In \emph{2023 IEEE International Conference on Image Processing (ICIP)}, pages 2815--2819. IEEE, 2023.

\bibitem[Zamir et~al.(2022)Zamir, Arora, Khan, Hayat, Khan, and Yang]{zamir2022restormer}
Syed~Waqas Zamir, Aditya Arora, Salman Khan, Munawar Hayat, Fahad~Shahbaz Khan, and Ming-Hsuan Yang.
\newblock Restormer: Efficient transformer for high-resolution image restoration.
\newblock In \emph{Proceedings of the IEEE/CVF conference on computer vision and pattern recognition}, pages 5728--5739, 2022.

\bibitem[Zhang et~al.(2022)Zhang, Li, Liang, Cao, Zhang, Tang, Timofte, and Van~Gool]{zhang2022practical}
Kai Zhang, Yawei Li, Jingyun Liang, Jiezhang Cao, Yulun Zhang, Hao Tang, Radu Timofte, and Luc Van~Gool.
\newblock Practical blind denoising via swin-conv-unet and data synthesis.
\newblock \emph{arXiv e-prints}, pages arXiv--2203, 2022.

\bibitem[Zhang et~al.(2023)Zhang, Lu, Chen, Wang, Shi, Wang, and Song]{zhang2023neural}
Yiwei Zhang, Guo Lu, Yunuo Chen, Shen Wang, Yibo Shi, Jing Wang, and Li Song.
\newblock Neural rate control for learned video compression.
\newblock In \emph{The Twelfth International Conference on Learning Representations}, 2023.

\bibitem[Zhu et~al.(2022)Zhu, Yang, and Cohen]{zhu2022transformer}
Yinhao Zhu, Yang Yang, and Taco Cohen.
\newblock Transformer-based transform coding.
\newblock In \emph{International Conference on Learning Representations}, 2022.

\bibitem[Zou et~al.(2022)Zou, Song, and Zhang]{zou2022devil}
Renjie Zou, Chunfeng Song, and Zhaoxiang Zhang.
\newblock The devil is in the details: Window-based attention for image compression.
\newblock In \emph{Proceedings of the IEEE/CVF conference on computer vision and pattern recognition}, pages 17492--17501, 2022.

\end{thebibliography}
}

\end{document}